\documentclass[letterpaper]{article} % DO NOT CHANGE THIS
\usepackage{aaai2026}  % DO NOT CHANGE THIS
\usepackage{times}  % DO NOT CHANGE THIS
\usepackage{helvet}  % DO NOT CHANGE THIS
\usepackage{courier}  % DO NOT CHANGE THIS
\usepackage[hyphens]{url}  % DO NOT CHANGE THIS
\usepackage{graphicx} % DO NOT CHANGE THIS
\usepackage{natbib}  % DO NOT CHANGE THIS AND DO NOT ADD ANY OPTIONS TO IT
\usepackage{caption} % DO NOT CHANGE THIS AND DO NOT ADD ANY OPTIONS TO IT
\usepackage{algorithm}
\usepackage{algorithmic}

\usepackage{newfloat}
\usepackage{listings}
\usepackage{amsmath} 
\usepackage{amssymb}
\usepackage{booktabs}
\usepackage{multirow}

\newcommand{\cmark}{\ding{51}} % check mark
\newcommand{\xmark}{\ding{55}} % cross mark
\usepackage{pifont}
\DeclareCaptionStyle{ruled}{labelfont=normalfont,labelsep=colon,strut=off} % DO NOT CHANGE THIS
\floatstyle{ruled}
\newfloat{listing}{tb}{lst}{}
\floatname{listing}{Listing}
\title{\raisebox{-0.3cm}{\includegraphics[width=1.7cm,height=1.3cm]{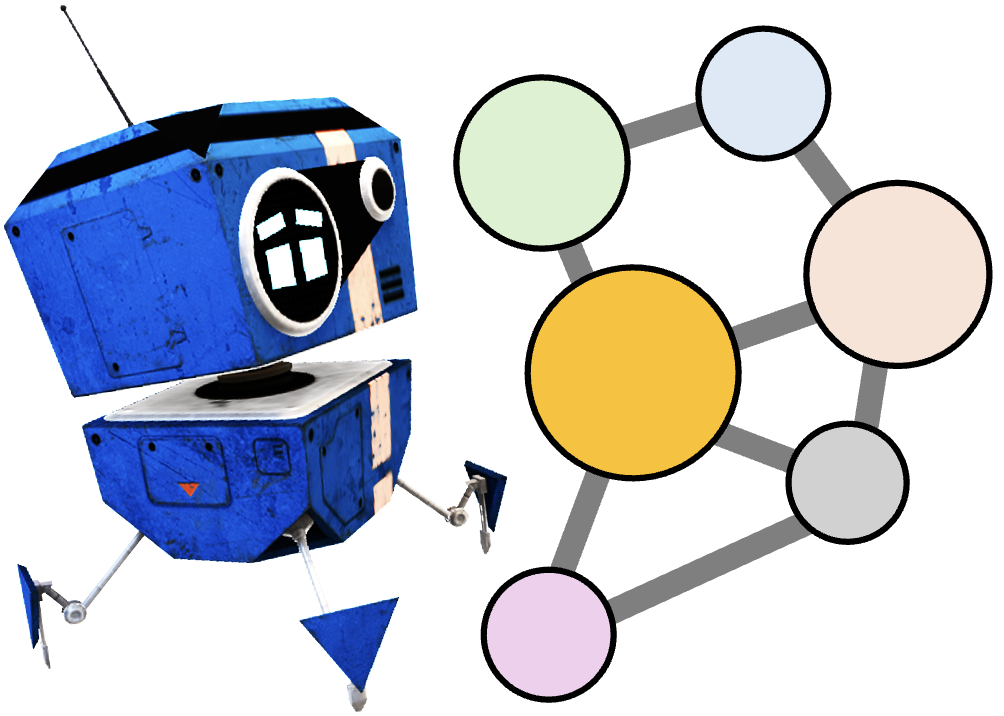}} View-on-Graph:\\Zero-Shot 3D Visual Grounding via Vision-Language Reasoning on Scene Graphs}
\author{
    Yuanyuan Liu\textsuperscript{\rm 1},
    Haiyang Mei\textsuperscript{\rm 1, \rm 2},\\
    Dongyang Zhan\textsuperscript{\rm 1},
    Jiayue Zhao\textsuperscript{\rm 1},
    Dongsheng Zhou\textsuperscript{\rm 3},
    Bo Dong\textsuperscript{\rm 4},
    Xin Yang\textsuperscript{\rm 1}\thanks{Corresponding Author}
}
\affiliations{
    \textsuperscript{\rm 1}Key Laboratory of Social Computing and Cognitive Intelligence, Dalian University of Technology\\
    \textsuperscript{\rm 2}Show Lab, National University of Singapore\\
    \textsuperscript{\rm 3}Dalian University\\
    \textsuperscript{\rm 4}Cephia AI\\
    liuyy990415@gmail.com, haiyang.mei@outlook.com, zhandongyang325@gmail.com, 2420458958@mail.dlut.edu.cn, zhoudongsheng@dlu.edu.cn, bo.dong@cephia.ai, xinyang@dlut.edu.cn
}

\begin{document}

\maketitle

\begin{abstract}
3D visual grounding (3DVG) identifies objects in 3D scenes from language descriptions. Existing zero-shot approaches leverage 2D vision–language models (VLMs) by converting 3D spatial information (SI) into forms amenable to VLM processing, typically as composite inputs such as specified-view renderings or video sequences with overlaid object markers. However, this \emph{VLM~$\oplus$~SI} paradigm yields entangled visual representations that compel the VLM to process entire cluttered cues, making it hard to exploit spatial–semantic relationships effectively.
In this work, we propose a new \emph{VLM~$\otimes$~SI} paradigm that externalizes the 3D SI into a form enabling the VLM to incrementally retrieve only what it needs during reasoning.
We instantiate this paradigm with a novel \textbf{View-on-Graph (VoG)} method, which organizes the scene into a multi-modal, multi-layer scene graph and allows the VLM to operate as an active agent that selectively accesses necessary cues as it traverses the scene. This design offers two intrinsic advantages:
(i) by structuring 3D context into a spatially and semantically coherent scene graph rather than confounding the VLM with densely entangled visual inputs, it lowers the VLM's reasoning difficulty; and
(ii) by actively exploring and reasoning over the scene graph, it naturally produces transparent, step-by-step traces for interpretable 3DVG.
Extensive experiments show that VoG achieves state-of-the-art zero-shot performance, establishing structured scene exploration as a promising strategy for advancing zero-shot 3DVG.
\end{abstract}

\begin{links}
    \link{Code}{https://github.com/YYLiuDLUT/VoG}
\end{links}

\section{Introduction}

3D visual grounding (3DVG) aims to localize objects in 3D scenes from natural language descriptions, a key capability for applications such as augmented reality~\cite{liu2023raydf, liu2025unleashing, wei2024sir}, vision–language navigation~\cite{chen2022think, gong2024cognition, huang2022assister}, and robotic perception~\cite{chen2023clip2scene, hu2024dhp, kong2023robo3d}. To solve this problem effectively, methods must integrate both textual understanding and spatial reasoning while coping with the complexity of 3D environments.

\begin{figure}[t]
    \centering
    \includegraphics[width=1\columnwidth]{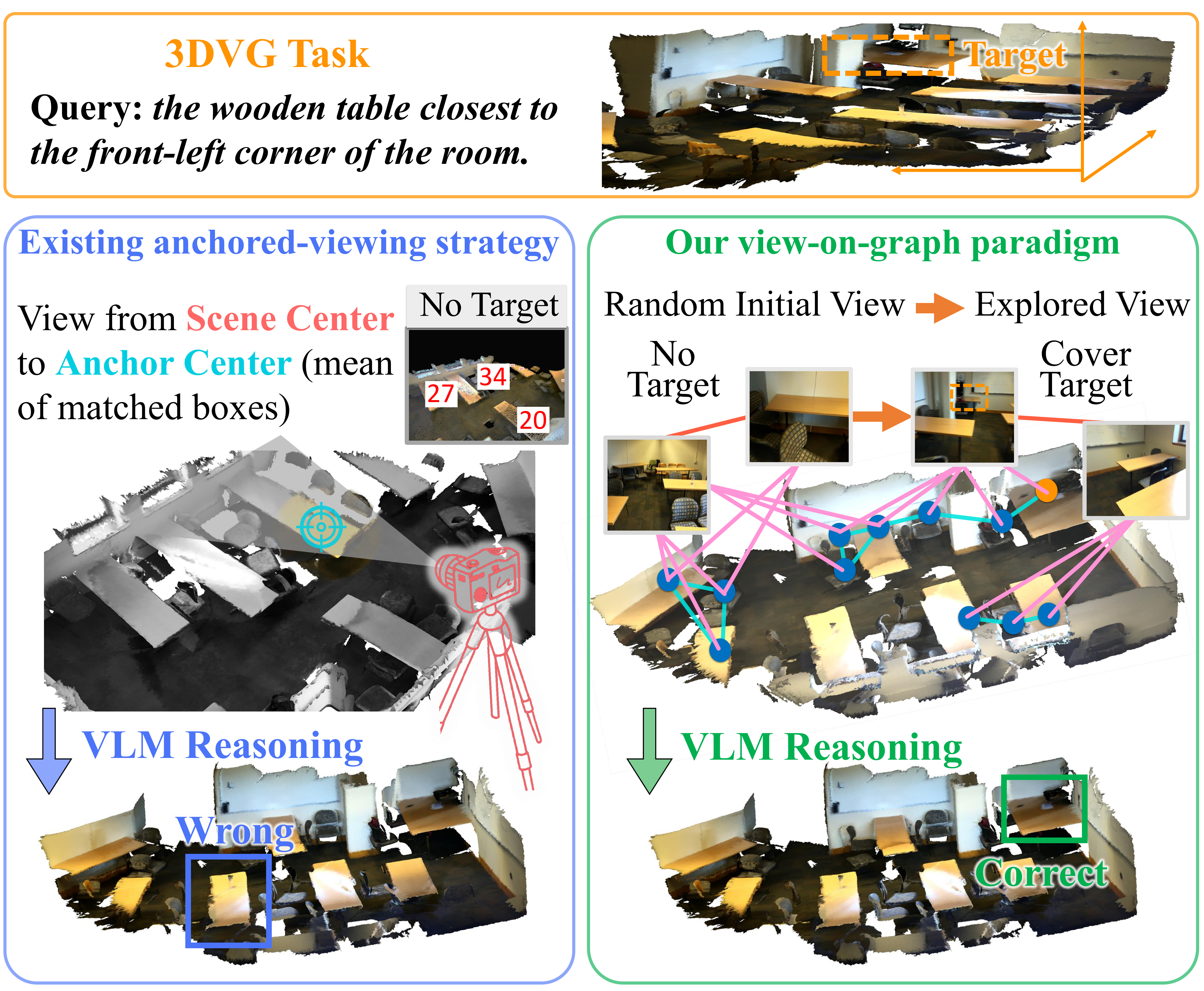}
    \caption{Comparison of zero-shot 3D visual grounding paradigms. \textbf{Left}: Passive, fixed-view processing paradigm. When the anchored-view observation misses the target, the VLM’s reasoning over incomplete or misleading visual cues leads to failure. \textbf{Right}: Active, iterative exploration via our View-on-Graph (VoG) paradigm. By traversing the scene graph to navigate from misleading views toward informative observations, the VLM accurately locates the target and produces interpretable grounding traces.}
    \label{fig:teaser}
\end{figure}

Recent work leverages 2D vision–language models (VLMs) for zero-shot 3DVG by converting 3D spatial information (SI) into forms amenable to VLM processing, typically as composite visual inputs such as specified-view renderings~\cite{li2025seeground} or video sequences~\cite{qi2025gpt4scene} with overlaid object markers. We refer to this as the \emph{VLM~$\oplus$~SI} paradigm, where the VLM passively consumes the entire entangled SI, which forces it to wade through cluttered cues and makes it hard to exploit spatial–semantic relationships effectively. This naturally raises a question: \textit{Can 3D spatial information be represented in a form that enables the VLM to incrementally retrieve only what it needs during its reasoning process?}

Much like how people use a search engine, we do not retain the entire internet in working memory. Instead, we retrieve relevant information step-by-step, progressively querying only what is necessary until the answer is found. Inspired by this principle of incremental retrieval, we design our zero-shot 3DVG approach to avoid overwhelming the VLM with the entire 3D SI upfront. Instead, we externalize the 3D SI into a structured, queryable scene graph (SG), enabling the VLM to operate as an active agent that selectively accesses spatial and semantic cues as it traverses the scene. This \emph{VLM~$\otimes$~SI} design inherently alleviates the reasoning difficulty associated with the entire entangled SI, maintains visual clarity, and supports more focused and effective reasoning.

To instantiate this paradigm, we design a novel \textbf{View-on-Graph (VoG)} method, which operates in two key stages. First, we transform the 3D spatial information into a multi-modal, multi-layer scene graph (MMMG) (Fig.~\ref{fig:teaser}). The MMMG consists of a view layer representing multi-view RGB images as nodes and an object layer representing detected 3D objects as nodes. Inter-layer edges connect views to the objects visible within them, while intra-layer edges link spatially adjacent viewpoints in the view layer and encode spatial relationships between objects in the object layer. Second, given a query, VoG enables the VLM to actively traverse this graph by alternating between exploration, where it moves along graph connections toward potentially informative viewpoints, and reasoning, where it cross-verifies observations and candidate objects from visited views. This iterative traversal continues until the target is confidently grounded or the search depth limit is reached, providing both targeted reasoning and interpretable grounding traces.

The advantage of this design is illustrated in Fig.~\ref{fig:teaser}. In our ``VLM~$\otimes$~SI'' paradigm (right), the VLM can actively traverse the scene graph to move from less informative views toward more informative ones, even when the target is not visible from the starting viewpoint. This traversal capability allows the model to progressively refine its reasoning and achieve accurate grounding in large or visually complex scenes. In contrast, the existing ``VLM~$\oplus$~SI'' paradigm (left) lacks such an exploration mechanism. When the anchored view misses critical visual evidence, the model is forced to reason over cluttered and potentially misleading cues, often leading to grounding failure.

In summary, this work makes the following contributions:
\begin{itemize}
    \item We propose a new VLM~$\otimes$~SI paradigm for zero-shot 3DVG, reframing the task as active and iterative scene exploration rather than passive processing of entire entangled 3D spatial information.
    \item We instantiate this paradigm with a novel View-on-Graph (VoG) method that structures 3D spatial information into a multi-modal, multi-layer scene graph and allows the VLM to autonomously traverse it during reasoning to identify critical cues for target grounding.
    \item We conduct extensive experiments validating the effectiveness of the VLM~$\otimes$~SI paradigm and showing that VoG achieves state-of-the-art zero-shot 3DVG performance, while inherently providing interpretable and traceable grounding through its step-by-step exploration.
\end{itemize}

\section{Related Work}
\textbf{3D Geometry-based Visual Grounding.} 3DVG localizes target objects in unstructured point clouds conditioned on language. Benchmarks such as ScanRefer~\cite{chen2020scanrefer} and ReferIt3D~\cite{achlioptas2020referit3d}. Early methods mainly rely on supervised training over geometric features and fall into two categories: two-stage pipelines~\cite{yuan2021instancerefer, cai20223djcg, zhao20213dvg}, which generate proposals\cite{guan2024gramo, mei2021exploring} before language matching, and single-stage pipelines~\cite{wu2023eda, wang2024g}, which directly align 3D points with text in an end-to-end manner. Two-stage methods offer modularity but sacrifice efficiency and generalization, while single-stage ones depend less on proposals but remain limited by sparse geometry and missing appearance cues.

\noindent\textbf{2D-enhanced Multi-modal Visual Grounding.} To mitigate the visual limitations of point clouds, another line incorporates 2D imagery as auxiliary input~\cite{zhang2024towards, bakr2022look}, enriching geometric context with high-resolution semantics. SAT~\cite{yang2021sat} projects multi-view RGB features into 3D space and uses attention-based fusion for language alignment. OpenScene~\cite{peng2023openscene} fuses dense 2D semantics with 3D voxels, enabling reasoning about color and material. 3DLFVG~\cite{zhang2024towards} transfers knowledge from 2D VLMs via cross-modal distillation, and LAR~\cite{bakr2022look} models relations between projected 2D features and 3D points. While effective, these methods demand extra modalities and annotations, raising scalability and training complexity concerns.

\begin{figure*}[t]
    \centering
    \includegraphics[width=.95\textwidth]{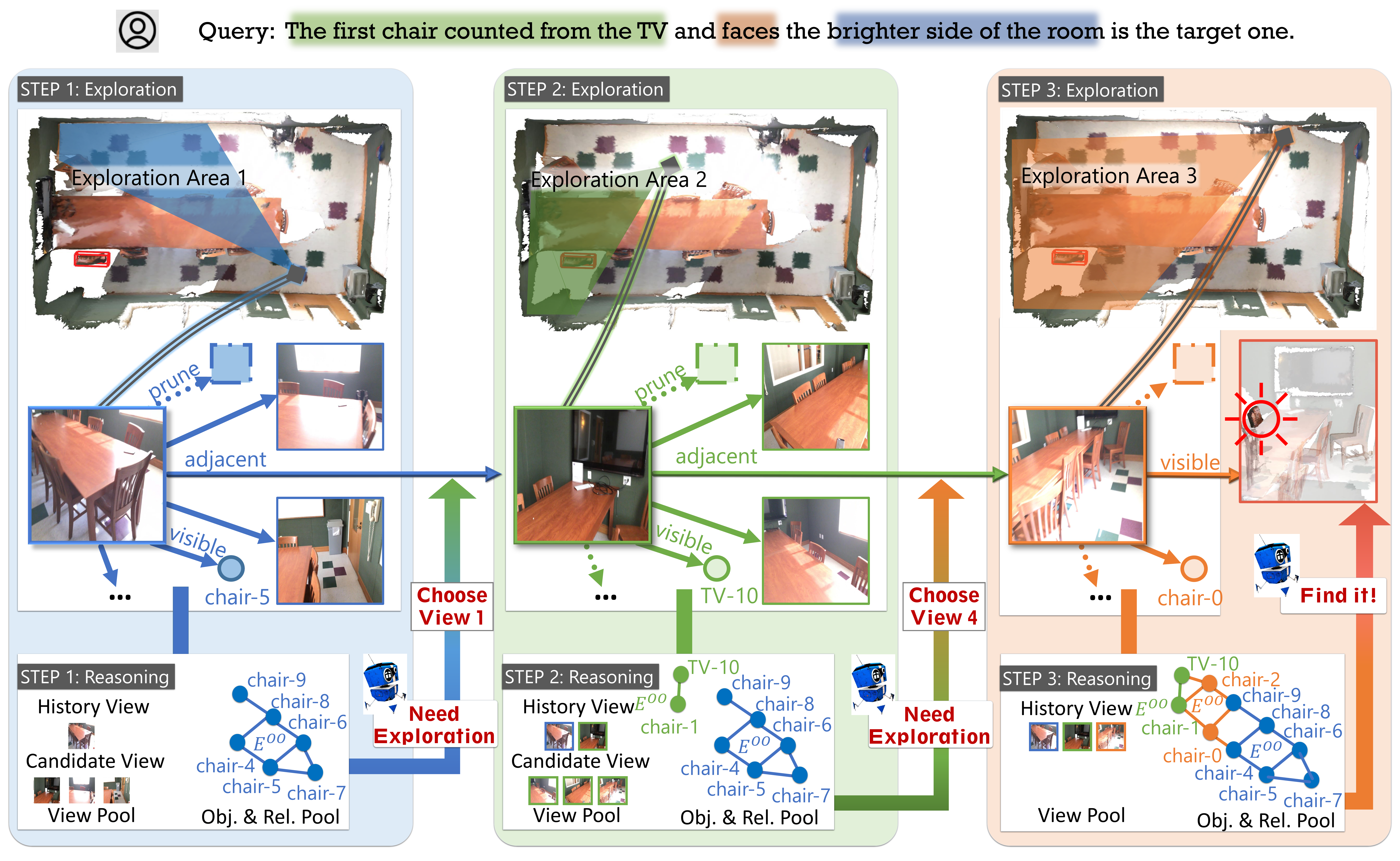}
    \caption{Workflow of VoG. Given the query, VoG identifies the target category (\texttt{chair}) and anchor (\texttt{TV}) as the search topic, then randomly selects an initial view where a chair is visible (View~0). From this view, it traverses neighboring nodes to form candidate nodes, which are fed to the VLM to decide whether further exploration is needed. The first exploration area reveals the ``brighter side of the room'' in the query, where multiple chairs are observed and added to the object pool. However, the anchor \texttt{TV} is missing, leaving the target unclear. The VLM then explores an area where the \texttt{TV} becomes visible (STEP~2) and observes ``the first chair counted from the TV,'' adding \texttt{chair-1} to the pool. Ambiguity remains due to another nearby chair, prompting a final exploration to capture the full spatial layout. Once all query cues, including ``faces the brighter side of the room'' are confirmed, the VLM identifies \texttt{chair-1} as the target and terminates the search.}
    \label{fig:pipeline}
\end{figure*}

\noindent\textbf{Zero-/Few-shot Visual Grounding Methods.} Recent trends shift toward few-shot~\cite{wang2023distilling} and zero-shot paradigms 3D-specific training~\cite{kang2024intent3d, zhu2024scanreason}, aiming to eliminate the dependency on large-scale 3D training data. These approaches leverage powerful VLMs pre-trained on 2D image-text pairs to perform 3D grounding via cross-modal reasoning. For example, SeeGround~\cite{li2025seeground} introduces a render-and-reason framework, where 2D renderings of 3D scenes are used to query VLMs for localization, enabling open-vocabulary generalization. Despite their strong generalization capabilities, these methods passively feed the VLM with the entire entangled 3D spatial information, forcing it to reason over cluttered cues and rely on implicit spatial understanding. In contrast, we externalize 3D SI into a structured, queryable SG, enabling the VLM to incrementally retrieve only the spatial–semantic cues it needs during traversal.

\section{Methodology}

We address zero-shot 3DVG by externalizing the 3D SI into a MMMG that separately encodes viewpoints, detected objects, and their spatial relations, rather than entangling them directly in the VLM’s inputs. The 3DVG task is then reformulated as a viewpoint-guided traversal over this graph, where the VLM alternates between exploring informative viewpoints and reasoning over the accumulated spatial–semantic evidence (Fig.~\ref{fig:pipeline}). The following sections detail the MMMG structure and the VoG framework.

\subsection{Multi-Modal \& Multi-Layered Scene Graph}
SGs are widely used in robotics and scene reasoning tasks for efficient navigation, planning, and multi-view integration~\cite{liu2022explore, rana2023sayplan, werby2024hierarchical, zhu2023combat, honerkamp2024language, koch2024sgrec3d, koch2024open3dsg, fang2025scalegraph, gao2024graphdreamer, sun2023think}.  
We extend this idea to 3DVG by proposing MMMG that compactly represents each scene as:
\begin{equation}
G = \{V, O, E^{VV}, E^{VO}, E^{OO}\},
\end{equation}
where \(V\) are view nodes with camera poses, \(O\) are object nodes with 3D boxes and semantics, and \(E^{VV}\), \(E^{VO}\), \(E^{OO}\) denote view–view connectivity, view–object visibility, and object–object spatial relations. These edges enable the VLM to reason across both visual and spatial modalities. The graph is built automatically from predicted 3D detections~\cite{schult2022mask3d} using point-cloud distances and camera poses. This layered design keeps visual and spatial information explicitly separated, avoiding modality entanglement while aligning 2D view observations with structured, queryable 3D knowledge. Further details are provided in the Appendix. As illustrated in Fig.~\ref{fig:pipeline} (STEP 1–3), each image corresponds to a node \(v\), which is linked to visible objects \(o\) via \(E^{VO}\) and connected to other views via \(E^{VV}\). The exploration area represents the visible region of \(v\).

\subsection{View-On-Graph}
Given the structured graph \( \mathcal{M_S} \) and query \(q\), VoG runs in three phases: initialization, exploration, and reasoning.

\textbf{Initialization.} Given a query, VoG leverages the underlying VLM to identify the target object \( O_T \) and candidate anchor objects \( \{ O_A \} \)~\cite{li2025seeground}. VoG then treats the \( O_T \) as the search topic and traverses the entire \( \mathcal{M_S} \) to collect all candidate views related to the topic. This yields an initial set of relevant views, from which one is randomly selected as the starting point. As illustrated in Fig.~\ref{fig:pipeline} (STEP 1), when the target is a \texttt{chair}, we randomly select the View 0 from the set of views in which chair is visible.

\textbf{Exploration.}  
At the beginning of the \( D \)-th iteration, each reasoning path \( P \) contains \( D-1 \) triples:
\begin{equation}
P_{D-1} = \{ (e_{d}^{sub}, r^{d}_{j}, e_{d}^{obj}) \}_{d=1}^{D-1},
\end{equation}
where \( e_{d}^{sub} \) and \( e_{d}^{obj} \) denote the subject and object nodes, respectively, and \( r^{d}_{j} \) represents the specific relation between them. \( e_{d}^{obj} \) and \( e_{d+1}^{sub} \) refer to the same node, forming \( P \) as a connected chain:
\(e_0 \rightarrow e_1 \rightarrow e_2 \rightarrow \dots \rightarrow e_d\).

In the \( D \)-th reasoning step, we start from the current endpoint \( e_{D-1} \) and aim to identify the relevant candidate nodes \( \mathcal{E}^{\text{cand}}_{D} \) among its neighbors with respect to the search topic.  
Since the underlying structure \( \mathcal{M_S} \) is a multi-modal and multi-layered graph, these candidate nodes may reside in different layers. Accordingly, exploration proceeds in two complementary modes:

\begin{itemize}
    \item \textbf{Intra-layer expansion}: Extend within the same layer (either view--view or object--object), filtering candidates based on semantic relevance to \( O_T \) and anchor objects \( \{ O_A \} \). This filtering preserves only nodes that contribute to the search topic, thus reducing ambiguity for the VLM:
    \begin{equation}
        \mathcal{E}^{\text{cand}}_{D_v} = \left\{ (e_{D-1}, r, e_D) \;\middle|\; r \in E^{VV},\; e_{D-1} \in \text{V} \right\}
    \end{equation}
    \begin{equation}
        \mathcal{E}^{\text{cand}}_{D_o} = \left\{ (e_{D-1}, r, e_D) \;\middle|\; r \in E^{OO},\; e_{D-1} \in \text{O} \right\}
    \end{equation}

    \item \textbf{Inter-layer transition}: Jump across layers from the current view node to its visible object nodes (no reverse), again filtering by semantic consistency:
    \begin{equation}
        \mathcal{E}^{\text{cand}}_{D_o'} = \left\{ (e_{D-1}, r, e_D) \;\middle|\; r \in E^{VO},\; e_{D-1} \in \text{V} \right\}
    \end{equation}
\end{itemize}

After this exploration phase, the combined set of filtered candidates is:
\begin{equation}
    \widetilde{\mathcal{E}}_D = \mathcal{E}^{\text{cand}}_{D_v} \cup \mathcal{E}^{\text{cand}}_{D_o} \cup \mathcal{E}^{\text{cand}}_{D_o'}
\end{equation}
To avoid redundant cycles and ensure forward progress, all nodes already visited in previous iterations are excluded from \(\widetilde{\mathcal{E}}_D\) before proceeding to the next reasoning step. As illustrated in Fig.~\ref{fig:pipeline} (Step~1), starting from View~0, we retain three relevant views containing chairs along with several visible chair object nodes. Irrelevant nodes are removed either due to sharing the same visual context or lacking semantic relevance to the topic. The remaining nodes form \(\widetilde{\mathcal{E}}_D\).

\textbf{Context Integration.} After each exploration step, we integrate the context to form a complete reasoning input for the VLM through two operations:
\begin{itemize}
    \item \textbf{Stitched View Construction:} We build an \( S\!\times\!S \) visual grid \( \text{I} \), placing previously explored views in the top-left and current candidate views in the remaining slots, with unused cells padded in white. This layout allows the VLM to visually compare historical and candidate views to assess redundancy or complementarity, and decide whether to continue view exploration or switch to object selection.
    \item \textbf{Accumulated Object Pool:} We maintain a growing pool of objects observed from explored views. These serve as global candidates, enabling the VLM to select the target object at any step and terminate reasoning early if confident.
\begin{equation}
    \mathcal{E}^{\text{cand}}_{D} = \left( \sum_{d=1}^{D-1} \mathcal{E}^{\text{cand}}_{d_o} \right) + \widetilde{\mathcal{E}}_D
\end{equation}
\end{itemize}

\textbf{Reasoning.} At each reasoning step, the VLM receives the grid image \(\text{I}\) and the candidate set \(\mathcal{E}^{\text{cand}}_{D}\). The system prompts the VLM to decide whether the current exploration area should be further expanded in the view space or transitioned to the object space. When a view candidate is chosen, the system repeats the ``Exploration'' and ``Reasoning'' steps, progressively refining and constraining the exploration area until it naturally converges to the object space or the maximum reasoning depth \( D_{\text{max}} \) is reached.  
If the depth limit is reached without object selection, a forced global reasoning step is performed over the accumulated multi-view and multi-object evidence.

As illustrated in Fig.~\ref{fig:pipeline}, VoG first explores the ``brighter side of the room'' finding multiple chairs but no \texttt{TV}, leaving the target ambiguous. It then moves to a view where the \texttt{TV} appears, spotting ``the first chair counted from the TV'' as \texttt{chair-1}. A nearby chair still causes uncertainty, so VoG explores once more to capture the full spatial layout. With all cues (``faces'') confirmed, \texttt{chair-1} is identified as the target and the process ends.

Overall, the VoG procedure consists of:
(i) an initial description parsing step;
(ii) iterative expansion/contraction of the exploration area over \( D \) reasoning rounds;
and (iii) a final global reasoning step if necessary — resulting in at most \( D + 2 \) VLM calls.

\section{Experiments}

\subsection{Experimental Settings}
\noindent\textbf{Datasets.}  
We evaluate on two widely used 3D visual grounding benchmarks.  
\textbf{ScanRefer}~\cite{chen2020scanrefer} contains 51,500 descriptions across 800 scenes with queries requiring either unique object localization or discrimination among same class distractors.  
\textbf{Nr3D}~\cite{achlioptas2020referit3d} comprises 41,503 precise descriptions, categorized into ``Easy'' or ``Hard'' depending on distractor count, and labeled as view-dependent or view-independent.  
These datasets cover both point-cloud only grounding and bounding-box based grounding scenarios, offering diverse and challenging settings to evaluate the spatial representational capacity of the VoG.

\begin{table*}[t]
  \centering
  \small
  \setlength{\tabcolsep}{1pt}
  \begin{tabular}{ll c  cc  cc  cc}
    \toprule
    \multirow{2}{*}{\textbf{Method}} 
    & \multirow{2}{*}{\textbf{Venue}}
    & \multirow{2}{*}{\textbf{Agent}}
    & \multicolumn{2}{c}{\textbf{Unique}}
    & \multicolumn{2}{c}{\textbf{Multiple}}
    & \multicolumn{2}{c}{\textbf{Overall}} \\
  \cmidrule(lr){4-5}\cmidrule(lr){6-7}\cmidrule(lr){8-9}
      &         &             
    & \textbf{Acc@0.25} & \textbf{Acc@0.5} 
    & \textbf{Acc@0.25} & \textbf{Acc@0.5} 
    & \textbf{Acc@0.25} & \textbf{Acc@0.5} \\
    \midrule
    \multicolumn{9}{l}{\textbf{\textit{Fully-Supervised}}} \\
    ScanRefer~\cite{chen2020scanrefer}      & ECCV’20 & —              & 67.6 & 46.2 & 32.1 & 21.3 & 39.0 & 26.1 \\
    InstanceRefer~\cite{yuan2021instancerefer}  & ICCV’21 & —              & 77.5 & 66.8 & 31.3 & 24.8 & 40.2 & 32.9 \\
    3DVG-T~\cite{zhao20213dvg}         & ICCV’21 & —              & 77.2 & 58.5 & 38.4 & 28.7 & 45.9 & 34.5 \\
    BUTD-DETR~\cite{jain2022bottom}      & ECCV’22 & —              & 84.2 & 66.3 & 46.6 & 35.1 & 52.2 & 39.8 \\
    EDA~\cite{wu2023eda}            & CVPR’23 & —              & 85.8 & 68.6 & 49.1 & 37.6 & 54.6 & 42.3 \\
    3D-VisTA~\cite{zhu20233d}       & ICCV’23 & —              & 81.6 & 75.1 & 43.7 & 39.1 & 50.6 & 45.8 \\
    G3-LQ~\cite{wang2024g}          & CVPR’24 & —              & 88.6 & 73.3 & 50.2 & 39.7 & 56.0 & 44.7 \\
    MCLN~\cite{qian2024multi}           & ECCV’24 & —              & 86.9 & 72.7 & 52.0 & 40.8 & 57.2 & 45.7 \\
    ConcreteNet~\cite{unal2024four}    & ECCV’24 & —              & 86.4 & 82.1 & 42.4 & 38.4 & 50.6 & 46.5 \\
    \midrule
    \multicolumn{9}{l}{\textbf{\textit{Weakly-Supervised}}} \\
    WS-3DVG~\cite{wang2023distilling}        & ICCV’23 & —              &  —   &  —   &  —   &  —   & 27.4 & 22.0 \\
    \midrule
    \multicolumn{9}{l}{\textbf{\textit{Zero-Shot}}} \\
    LERF~\cite{kerr2023lerf}           & ICCV’23 & CLIP          &  —   &  —   &  —   &  —   &  4.8 &  0.9 \\
    OpenScene~\cite{peng2023openscene}      & CVPR’23 & CLIP          & 20.1 & 13.1 & 11.1 &  4.4 & 13.2 &  6.5 \\
    LLM-G~\cite{yang2024llm}          & ICRA’24 & GPT-3.5       &  —   &  —   &  —   &  —   & 14.3 &  4.7 \\
    LLM-G~\cite{yang2024llm}          & ICRA’24 & GPT-4 Turbo   &  —   &  —   &  —   &  —   & 17.1 &  5.3 \\
    ZSVG3D~\cite{yuan2024visual}         & CVPR’24 & GPT-4 Turbo   & 63.8 & 58.4 & 27.7 & 24.6 & 36.4 & 32.7 \\
    SeeGround~\cite{li2025seeground}      & CVPR'25 & Qwen2-VL-2B   & 59.9 & 55.0 & 20.7 & 18.4 & 30.2 & 27.2 \\
    SeeGround~\cite{li2025seeground}      & CVPR'25 & Qwen2-VL-72B  & 75.7 & 68.9 & \bfseries 34.0 & 30.0 & 44.1 & 39.4 \\
    \midrule
    \textbf{VoG}            & \textbf{Ours}    & Qwen2-VL-2B   & 69.1 & 63.1 & 25.3 & 21.9 & 35.9 & 31.9 \\
    \textbf{VoG}            & \textbf{Ours}    & Qwen2-VL-72B  & \bfseries 78.6 & \bfseries 71.5 & \bfseries 34.0 & \bfseries 30.4 & \bfseries 44.8 & \bfseries 40.3 \\
    \bottomrule
  \end{tabular}
  % \vspace{-4pt}
  \caption{Evaluations of 3DVG on \textit{ScanRefer} validation set. Results are reported for “Unique” (scenes with a single target object) and “Multiple” (scenes with distractors of the same class) subsets, along with overall performance.}
  \label{tab:scanrefer_results}
\end{table*}

\noindent\textbf{Implementation Details.}   
We adopt Qwen2-VL-2B~\cite{wang2024qwen2} as the VLM. 
The MMMG viewpoint layer is built from ScanNet RGB images,  using K-Means clustering camera poses into \(M = 16\) representative views, with a maximum reasoning depth \(D_{\text{max}} = 4\). All experiments are zero-shot on 8$\times$~NVIDIA H200 GPUs, 
yielding an average latency of 0.27\,s/query (2.17\,s/query per GPU) for Qwen2-VL-2B and 
1.51\,s/query (12.10\,s/query per GPU) for Qwen2-VL-72B. Further details are provided in the Appendix.

\subsection{Comparative Study}

\noindent\textbf{ScanRefer.}  
Tab.~\ref{tab:scanrefer_results} shows that our VoG framework sets a new state-of-the-art in the \emph{zero-shot} setting.  
\textbf{(1) Large model advantage:} VoG-72B surpasses all existing zero-shot methods across all metrics, and even outperforms the weakly-supervised WS-3DVG.  
\textbf{(2) Small model efficiency:} VoG-2B, despite being \textbf{36$\times$ smaller} than SeeGround-72B, achieves \textbf{81.4\%} of its Acc@0.25 performance and outperforms most prior zero-shot baselines.  
\textbf{(3) Competitiveness with supervision:} While trained with no 3DVG annotations, VoG-72B achieves results on par with several fully-supervised methods.

\noindent\textbf{Nr3D.}  
Tab.~\ref{tab:nr3d_results} further confirms our findings on this complementary benchmark.  
VoG-72B consistently leads all zero-shot baselines across \emph{Easy}, \emph{Hard}, \emph{Dep.}, and \emph{Indep.} splits, while narrowing the gap to fully-supervised methods~\cite{achlioptas2020referit3d, chen2022language, huang2021text, chang2024mikasa}.Meanwhile, VoG-2B maintains strong zero-shot competitiveness despite its compact size.

\subsection{Visual Grounding Traceability}

\begin{figure}[htbp]
    \centering
    \includegraphics[width=0.9\columnwidth]{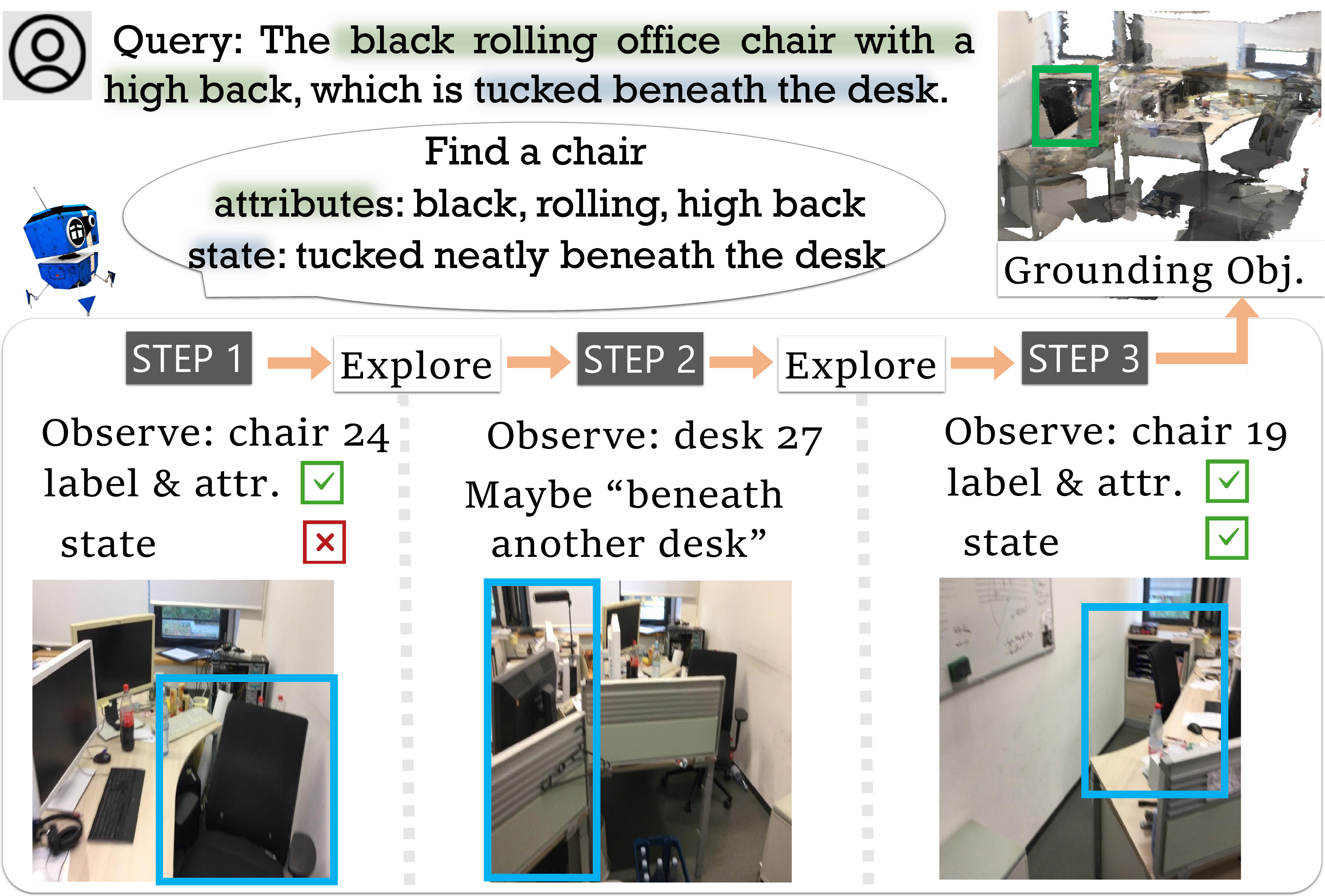}
    \caption{Traceable reasoning paths of VoG. Starting from an initial view showing a visually similar but mismatched chair, VoG rejects it due to missing the ``tucked beneath the desk'' cue. It then explores related scene elements (\texttt{desk}) and progressively refines its hypothesis. The search converges when both appearance and spatial state match, resulting in correct grounding.}
    \label{fig:tracibility}
\end{figure}

A unique advantage of VoG is its fully \textbf{traceable} reasoning process. Rather than outputting only the final grounded object, VoG reveals its sequence of exploratory steps. As shown in Fig.~\ref{fig:tracibility}, the initial view contains a confusing chair with similar appearance (“black, rolling, high back”), yet VoG refrains from prematurely selecting it, showing that it actively aligns visual cues with the textual description (e.g., rejects mismatches such as ``tucked beneath the desk''). It then shifts attention to related scene elements (e.g., nearby desks) and refines its hypothesis before converging on the correct object. This coherent, interpretable search path turns VoG from a ``black-box'' predictor into a transparent decision-maker, facilitating error diagnosis, trust, and insight into how appearance, attributes, and spatial relations guide grounding.

\begin{table}[tbp]
  \centering
  \small
  \setlength{\tabcolsep}{2.3pt}
  \begin{tabular}{l l c c c c c}
    \toprule
    \textbf{Method} & \textbf{Venue} & \textbf{Easy} & \textbf{Hard} & \textbf{Dep.} & \textbf{Indep.} & \textbf{Overall} \\
    \midrule
    \multicolumn{7}{l}{\textbf{\textit{Fully-Supervised}}} \\
    ReferIt3DNet   & ECCV'20        & 43.6 & 27.9 & 32.5 & 37.1 & 35.6 \\
    TGNN           & AAAI'21        & 44.2 & 30.6 & 35.8 & 38.0 & 37.3 \\
    InstanceRefer  & ICCV'21        & 46.0 & 31.8 & 34.5 & 41.9 & 38.8 \\
    3DVG-T         & ICCV'21  & 48.5 & 34.8 & 34.8 & 43.7 & 40.8 \\
    BUTD-DETR      & ECCV'22        & 60.7 & 48.4 & 46.0 & 58.0 & 54.6 \\
    MiKASA         & CVPR'24        & 69.7 & 59.4 & 65.4 & 64.0 & 64.4 \\
    ViL3DRel       & —        & 70.2 & 57.4 & 62.0 & 64.5 & 64.4 \\
    \midrule
    \multicolumn{7}{l}{\textbf{\textit{Weakly-Supervised}}} \\
    WS-3DVG        & ICCV'23  & 27.3 & 18.0 & 21.6 & 22.9 & 22.5 \\
    \midrule
    \multicolumn{7}{l}{\textbf{\textit{Zero-Shot}}} \\
    ZSVG3D-GPT4         & CVPR'24  & 46.5 & 31.7 & 36.8 & 40.0 & 39.0 \\
    SeeGround-2B   & CVPR'25  & 31.8 & 16.6 & 22.8 & 24.5 & 23.9 \\
    SeeGround-72B  & CVPR'25  & 51.3 & 35.6 & 38.8 & 45.5 & 43.1 \\
    \midrule
    \textbf{VoG-2B}  & \textbf{Ours} & 40.3 & 22.6 & 30.4 & 31.5 & 31.1 \\
    \textbf{VoG-72B} & \textbf{Ours} & \textbf{58.9} & \textbf{37.2} & \textbf{39.4} & \textbf{52.1} & \textbf{47.6} \\
    \bottomrule
  \end{tabular}
  % \vspace{-5pt}
  \caption{Performance on Nr3D validation set. Queries are labeled as \emph{Easy} (one distractor) or \emph{Hard} (multiple distractors), and as \emph{View-Dependent} or \emph{View-Independent} based on viewpoint requirements for grounding.}
  \label{tab:nr3d_results}
\end{table}

\begin{figure}[htbp]
    \centering
    \includegraphics[width=0.9\columnwidth]{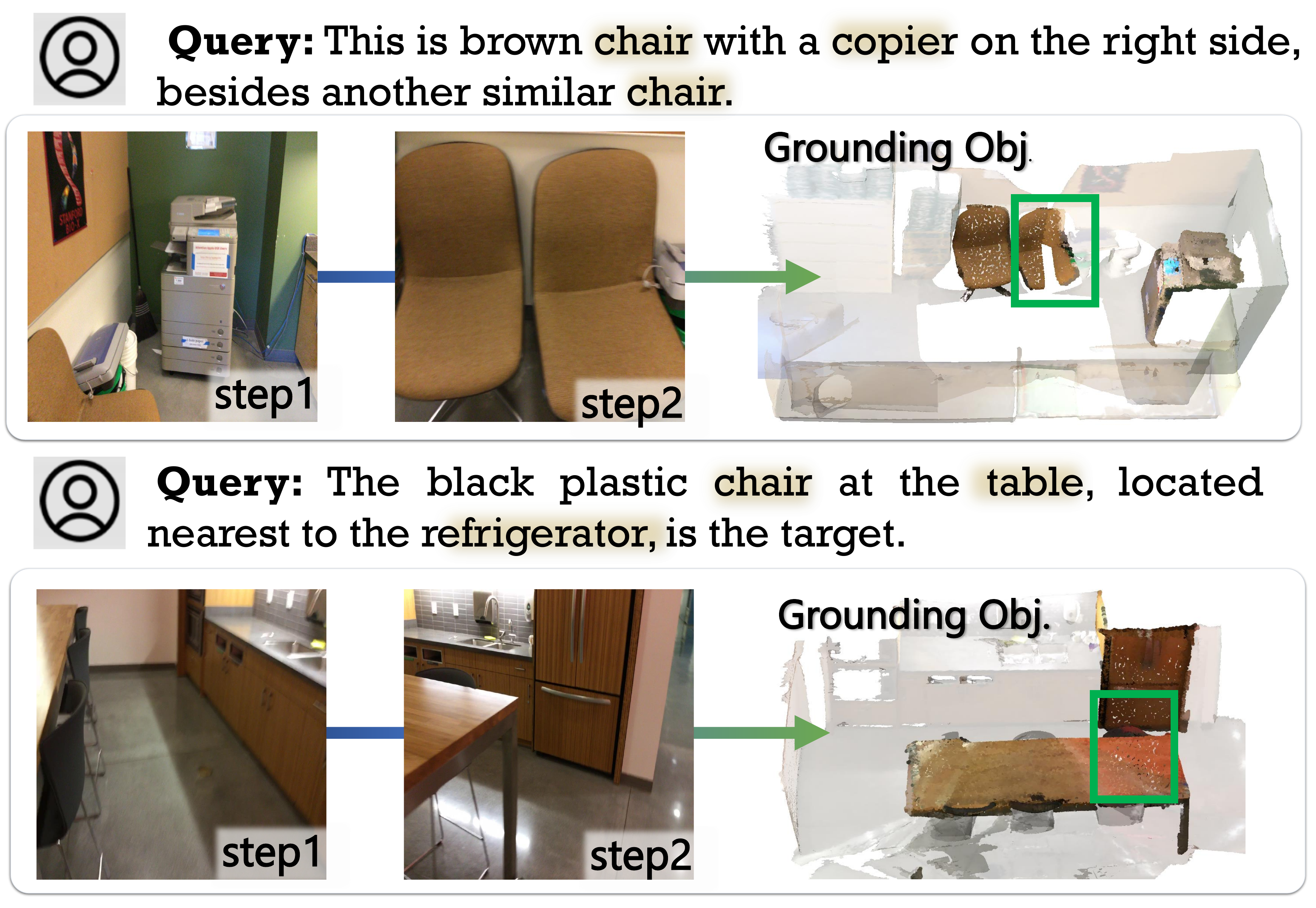}
    \caption{Qualitative grounding results. (\textbf{Top}) The initial view reveals a brown chair matching part of the query description (``copier machine to the right''), but VoG explores further to verify spatial cues (“beside another similar chair”) before confirming the target. (\textbf{Bottom}) Starting from ``at the table'', VoG explores to confirm ``nearest to the refrigerator'' before final grounding.}
    \label{fig:visualization}
\end{figure}

\subsection{Ablation Study}
We perform various ablation studies to understand the importance of different factors in VoG. We conduct our ablation studies on ScanRefer, use the open-source Qwen2-VL-2B as the VLM, and the performance is measured using overall Acc@0.25.

\begin{figure*}[tbp]
    \centering
    \includegraphics[width=1\textwidth]{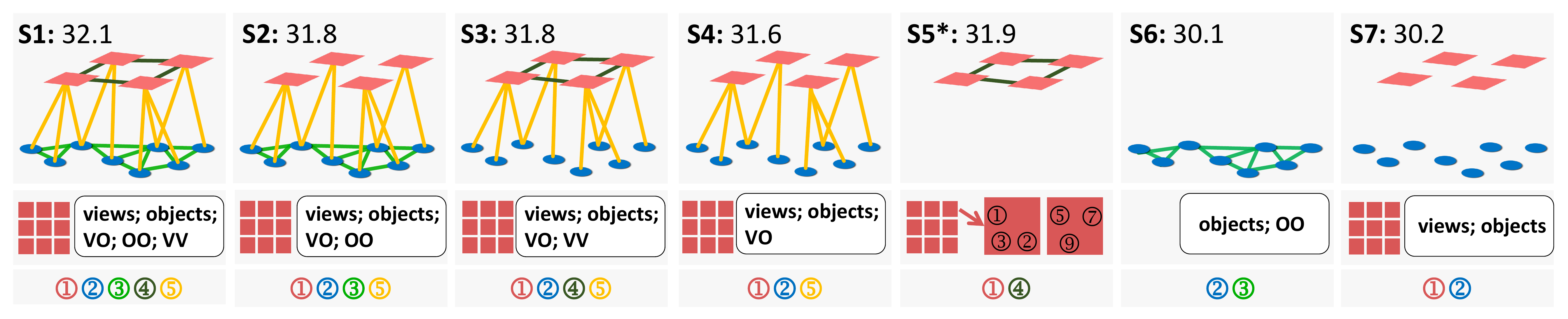}
    \caption{ Graph Structure Ablation.  S1: Full structure. S2–S4: Remove one type of edge while keeping others.  S5*: Keep only images input with global object IDs.   S6: Keep only text input.  S7: Remove all structure.  The first row shows the graph structure configurations and the corresponding accuracy.
}

    \label{fig:ablation1}
\end{figure*}

\noindent\textbf{(1) Do different Tree Structures affect 3DVG's performance?}  
We study the impact of MMMG's structural components by disabling multi-round reasoning and feeding all information in a single round. As shown in Fig.~\ref{fig:ablation1}, removing any component consistently harms performance, with the absence of all structure (S7) causing the largest drop, confirming that MMMG is crucial for spatial reasoning. Note that in S5*, we annotate each image with a global ID, same as~\cite{qi2025gpt4scene}, to assist the VLM in identifying the objects. This global ID marker serves as a form of View-Object relations, so the conclusions drawn from this configuration should be considered with reservation.

\begin{table}[tbp]
\centering
\small
\renewcommand{\arraystretch}{1.1}
\setlength{\tabcolsep}{6pt}
\begin{tabular}{cccccc}
\toprule
\textbf{Variants} & \textbf{S} & \textbf{Round}  & \textbf{Graph Traversal} &\textbf{Pool} & \textbf{Acc} \\
\midrule
R1(ours) & S1  & \cmark & \cmark & \cmark & 35.9 \\
R2 & S1  & \cmark & \xmark & \cmark  &30.3 \\
R3 & S1  & \cmark & \cmark & \xmark  &32.7 \\
R4 & S1  & \cmark & \xmark  & \xmark  &32.6 \\
R5 & S1  & \xmark & \xmark  & N/A  &32.1 \\
R6 & S4  & \cmark & N/A  & \cmark & 27.9 \\
R7 & S4  & \cmark & N/A  & \xmark & 31.7 \\
R8 & S4  & \xmark & N/A & N/A  &31.6 \\
\bottomrule
\end{tabular}
\caption{Ablation study on the VoG modules. \textbf{S} denotes the scene structure used in Fig.~\ref{fig:ablation1}, \textbf{Round} indicates multi-step reasoning. \textbf{Pool} denotes context integration.}
\label{tab:iteration}
\end{table}

\begin{figure*}[t]
    \centering
    \includegraphics[width=1\textwidth]{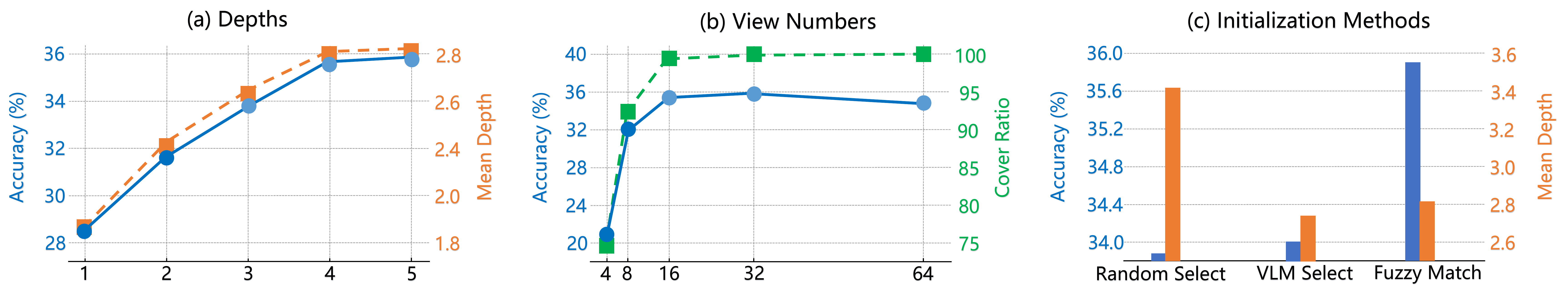}
    \caption{Ablation on reasoning depth, view number, and initialization methods.}
    \label{fig:ablation2}
\end{figure*}

\noindent\textbf{(2) How well do VoG modules complement each other?}  
We analyze two fixed structures, Full Structure (S1) and Minimal Structure (S4), to evaluate the contribution of each module (Tab.~\ref{tab:iteration}). To implement multi-round reasoning without graph traversal and pool (R4 and R7), we randomly select $M/\text{max\_step}$ view nodes in each round to ensure all information is seen within the \(D_{max}\) steps. Note that, N/A means modules cannot be performed.
\begin{itemize}
    \item Graph Traversal Effects: By filtering key information in each round, graph traversal prevents VoG from over-exploring irrelevant regions (R1 vs R2). Without it, noisy candidates are repeatedly explored, causing most reasoning paths to fail even reaching \(D_{max}\) (R2 vs R4, R6 vs R7);
    \item Context Integration Effects: Enabling VoG to reuse promising candidates rather than starting from scratch each round (R1 vs R3). Without it, well-filtered candidates from graph traversal are discarded after each round, leading to the loss of valuable exploration history context.
    \item Complementarity of Graph Traversal \& Context Integration: When combined both modules (R1), graph traversal ensures that only high-quality candidates are preserved, while context integration carries them forward across rounds. This combination enables coherent, focused exploration throughout reasoning (Fig.~\ref{fig:visualization}).
    \item Multi-round Reasoning vs. One-round: Multi-round reasoning remains more effective than providing all information at once, as it allows incremental retrieval and reasoning at each step (R4 vs R5, R7 vs R8).
\end{itemize}

\noindent\textbf{(3) How does search depth affect VoG?}  
We evaluate \texttt{max\_step} from 1 to 5 (Fig.~\ref{fig:ablation2} (a)). Accuracy improves with depth, but the gain saturates beyond 4 since most questions require reasoning depths no greater than 3. Considering the linear cost growth, we set 4 as the max step.

\noindent\textbf{(4) How does graph size affect VoG?}  
The MMMG size depends on the number of view and object nodes. Since all objects are included, we focus on view nodes. Too few views yield low coverage and poor accuracy (Fig.~\ref{fig:ablation2} (b)). Increasing views improves both metrics, but beyond 32, excessive candidates per round make selection harder and slightly degrade performance. In VoG, the view images set as 16.

\noindent\textbf{(5) How does path initialization affect VoG?}  
The initial view determines VoG’s starting point. The closer it is to the target, the fewer reasoning steps are needed.  Our method selects it from fuzzy-matching the target and anchor in the VLM-processed description. Compared with random selection and direct VLM choice. As shown in Fig.~\ref{fig:ablation2} (c), our method shows a certain degree of error correction, achieving competitive results even with random initialization. Direct VLM selection causes a significant accuracy drop, most searches stop at step~1 because the chosen view already contains most query-related information, discouraging further exploration and causing frequent misclassification.

\begin{figure}[t]
    \centering
    \includegraphics[width=1\columnwidth]{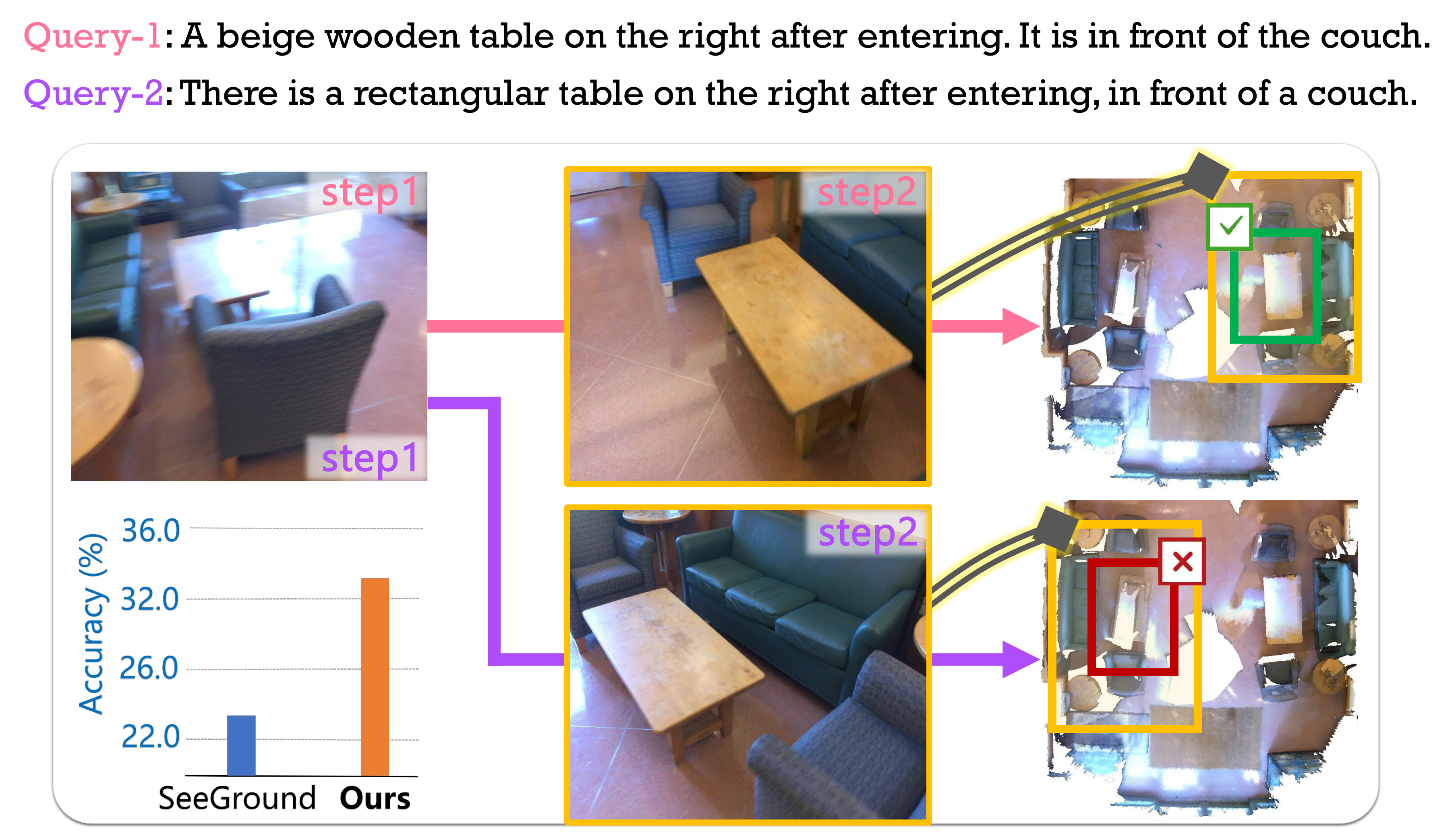}
    \caption{VoG efficacy drops in axis-symmetric scenes where Query 2 refers to the same target as Query 1, but in Step 2 selects a visually similar yet opposite-side view, resulting in grounding failure.}
    \label{fig:limitation}
\end{figure}

\section{Limitations}
VoG struggles in fully axis-symmetric scenes. In the example of Fig.~\ref{fig:limitation}, although \textit{Query~2} refers to the same target as \textit{Query~1}, VoG selects a visually similar but opposite-side view at Step~2, ultimately leading to grounding failure. Such symmetric layouts are inherently challenging for existing methods. In this scene, VoG achieves an average accuracy of \(32.8\%\) whereas SeeGround \cite{li2025seeground} attains \(23.4\%\). This suggests a future work of enhancing the scene graph with stronger spatial encoding, such as geometric constraints or pose-aware relational edges, to better preserve spatial continuity and resolve symmetric ambiguities.

\section{Conclusion}
We introduced View-on-Graph (VoG), a framework that redefines zero-shot 3DVG from the existing \emph{VLM~$\oplus$~SI} formulation to an interactive \emph{VLM~$\otimes$~SI} paradigm. By structuring 3D spatial information into a multi-modal, multi-layer scene graph and enabling the VLM to perform active and interactive exploration, VoG alleviates reasoning difficulty and enhances interpretability through step-by-step traceable grounding. Extensive experiments demonstrate the effectiveness of this \emph{VLM~$\otimes$~SI} paradigm and show that VoG achieves state-of-the-art zero-shot 3DVG performance, providing clear evidence that explicit scene structuring and interactive exploration can advance zero-shot 3DVG.

\begin{center}
    {\Large \textbf{Supplementary Material}}\\[0.8cm]
\end{center}

\section*{Overview}
The supplementary material includes the following contents:

\begin{table}[h]
\centering
\setlength{\tabcolsep}{16pt}
\begin{tabular}{c l} % 第一列居中
\toprule
\textbf{Section} & \textbf{Description} \\
\midrule
\multirow{1}{*}{Appendix A\ref{app:a}} 
& MMMG construction \\
\midrule
\multirow{2}{*}{Appendix B} 
& VLM prompt design \\
& VoG candidates generation \\
\midrule
\multirow{2}{*}{Appendix C} 
& Different VLM agents\\
& Graph structure analyses \\
\midrule
\multirow{1}{*}{Appendix D} 
& Qualitative grounding results \\
\midrule
\multirow{1}{*}{Appendix E} 
& Future work \\
\bottomrule
\end{tabular}
\end{table}

\begin{figure*}[t]
    \centering
    \includegraphics[width=1\textwidth]{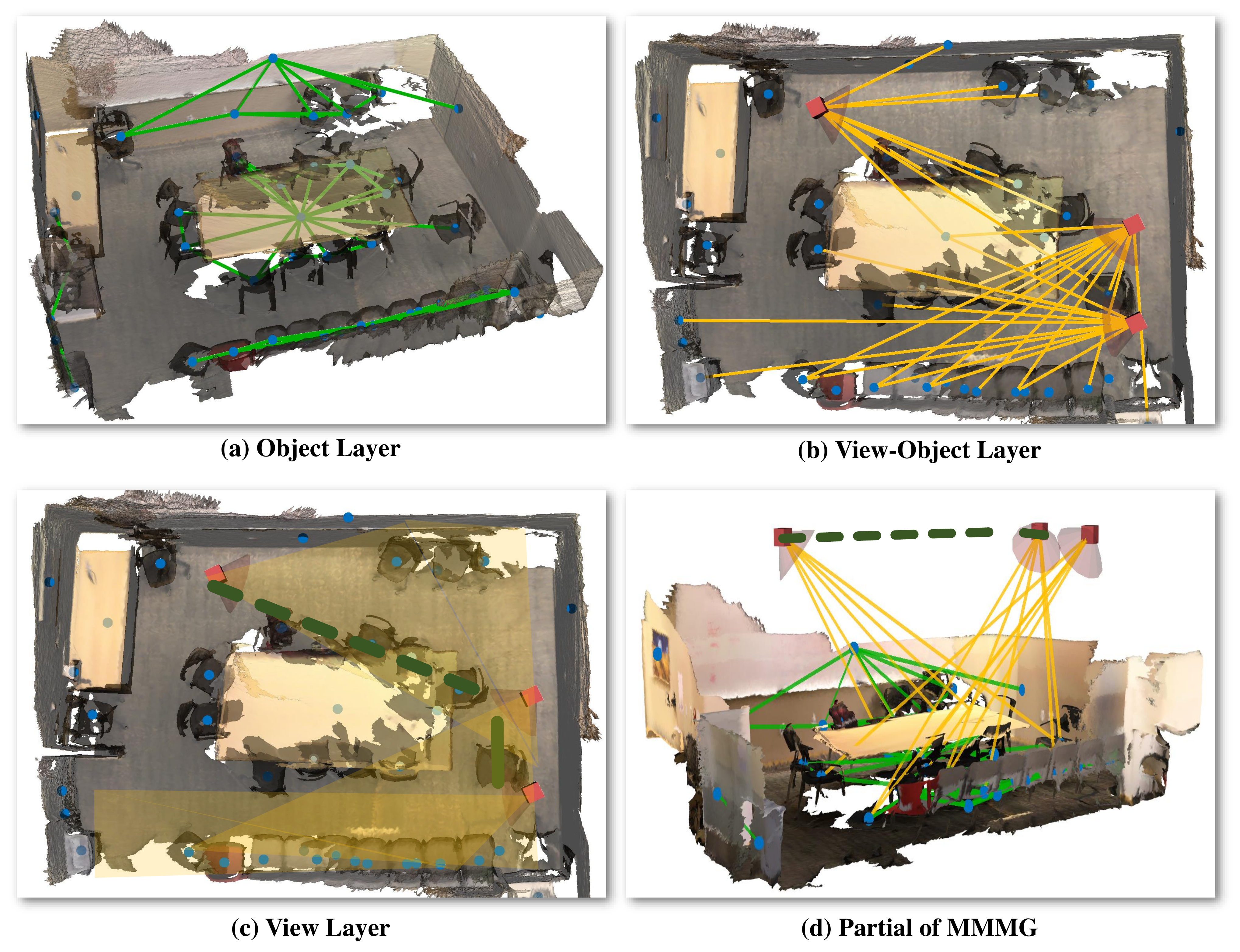}
    \caption{ Illustration of the MMMG structure. Blue circles denote $O$, red squares denote $V$. Green lines represent $E^{OO}$. Yellow lines represent $E^{VO}$. Dark green lines represent $E^{VV}$, where solid lines indicate \textit{adjacent} edges (moderate object overlap) and dashed lines indicate \textit{complementary} edges (minimal overlap). The yellow shaded area depicts the visible field of view for a given camera, highlighting its coverage of the scene.}
    \label{fig:graph}
\end{figure*}

\section{Appendix A}\label{app:a}

\subsection{Multi-Modal \& Multi-layered Scene Graph} 

In the main paper, we described the structure of the proposed Multi-Modal, Multi-Layered Scene Graph (MMMG) and its role in enabling active and iterative 3DVG reasoning.  
Here, we provide the detailed construction process of MMMG to complement the methodology in the main text.

3D Scene Graphs (3DSGs) were originally proposed to represent structured scene information, capturing both object attributes and their spatial arrangement. Building on this concept, we extend 3DSGs by introducing a Multi-Modal, Multi-Layered Graph structure. Unlike traditional 3DSGs, we represent each 3D scene as a graph:
\begin{equation}
G = \{V, O, E^{VV}, E^{VO}, E^{OO}\},
\end{equation}
where \(V\) and \(O\) denote the sets of view and object nodes, respectively. \( V = \{ v_1, v_2, \dots, v_M \} \), \(M = |V|\) is the number of camera views, and each \(v_i\) is associated with a corresponding camera pose \(CP_i\) (Fig.~\ref{fig:graph}, red square). The object node set \(O\) includes all objects detected in the 3D scene \(S\). Each object node is defined as \( o_i = \{\text{bbox}_i, \text{sem}_i\} \), where \(\text{bbox}_i\) represents the 3D bounding box (center and size), and \(\text{sem}_i\) denotes the semantic label (Fig.~\ref{fig:graph}, blue node).

Based on \(V\) and \(O\), we define three types of edges:

\paragraph{1. Object–Object Edges (\(E^{OO}\)).}
These edges model spatial relationships (\texttt{left}, \texttt{right}, \texttt{front}, \texttt{behind}, \texttt{above}, \texttt{below}) between object nodes, forming the object layer of MMMG:
\(\text{Layer}_1 = \{O, E^{OO}\}\).
To determine whether two object nodes \( (o_i, o_j) \) are connected, we first filter neighbors using a radius-based search to avoid noisy long-range connections:
\begin{equation}
    \text{E}^{OO} = \{(o_i, o_j) \mid o_j \in \text{N}(o_i)\}
\end{equation}
\begin{equation}
    \text{N}(o_i) = \left\{ o_j \mid o_j \in O,\, o_j \ne o_i,\, \text{Con}(o_i, o_j) \right\}
\end{equation}
\begin{equation}
    \text{Con}(o_i, o_j) = \exists\, \mathbf{p} \in o_i,\, \mathbf{q} \in o_j:\; \|\mathbf{p} - \mathbf{q}\| < r,
\end{equation}
where \(\mathbf{p}\) and \(\mathbf{q}\) are points belonging to \(o_i\) and \(o_j\), respectively. \(\text{Con}(o_i, o_j)\) holds if there exists at least one point pair within distance threshold \(r\). For each adjacent object pair in \(\text{N}(o_i)\), we assign a directional spatial relation based on their relative position. These edges encode local spatial layout among objects, enabling fine-grained geometric reasoning (Fig.~\ref{fig:graph}, green lines).

\paragraph{2. View–Object Edges (\(E^{VO}\)).}
These edges connect view nodes and object nodes, linking the two layers of MMMG:
\(\text{Layer}_{1,2} = \{V, O, E^{VO}\}\).
A connection \((v_i, o_j)\) is established if object \(o_j\) is visible from view \(v_i\):
\begin{equation}
    E^{VO} = \left\{ (v_i, o_j) \mid o_j \in \text{Vis}(v_i) \right\}.
\end{equation}
We project object point clouds onto the image plane of \(v_i\) using intrinsic matrix \(K\) and extrinsic parameters \((R_c, T_c)\). A depth map records the nearest depth per pixel for visibility reasoning. An object is visible if a sufficient portion of its projected points is unoccluded. \(E^{VO}\) is crucial for aligning image-level content with 3D object nodes (Fig.~\ref{fig:graph}, yellow lines).

\paragraph{3. View–View Edges (\(E^{VV}\)).}
These edges capture visual relationships between views, forming the view layer of MMMG:
\(\text{Layer}_2 = \{V, E^{VV}\}\).
Given two views \(v_i\) and \(v_j\), we compute their visible object sets \(\mathcal{O}_i\) and \(\mathcal{O}_j\), then measure their Intersection-over-Union (IoU). Using thresholds \(\tau_{\text{low}}\) and \(\tau_{\text{high}}\), we categorize edges as:
\begin{equation}
    E^{VV}_{ij} =
    \begin{cases}
        \text{complementary}, & \text{IoU} < \tau_{\text{low}}, \\
        \text{adjacent}, & \tau_{\text{low}} \le \text{IoU} < \tau_{\text{high}}, \\
        \text{(no edge)}, & \text{otherwise}.
    \end{cases}
\end{equation}
\(E^{VV}\) promotes exploration of both overlapping and complementary views, ensuring balanced contextual coverage (Fig.~\ref{fig:graph}, dark green lines).

Through this construction, each scene yields a unique MMMG representation \(\mathcal{M_S}\), shared across all descriptions in that scene, serving as the structured backbone for our VoG reasoning process.

\begin{figure*}[t]
\centering
\includegraphics[width=.95\textwidth]{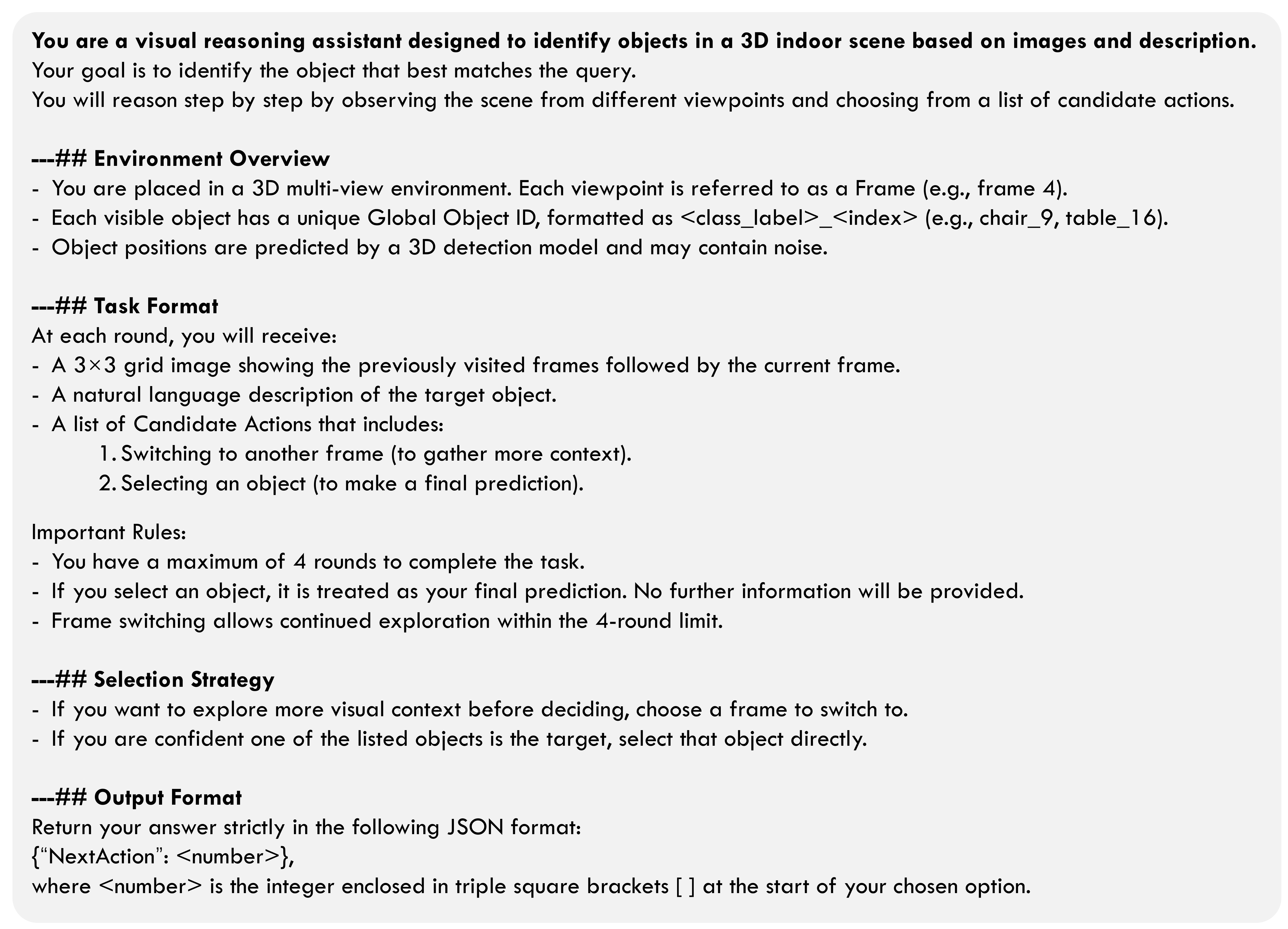}
\caption{The instruction used for prompting the VLMs to give a decision in each step.}
\label{fig:prompt}
\end{figure*}

\begin{figure*}[t]
\centering
\includegraphics[width=.95\textwidth]{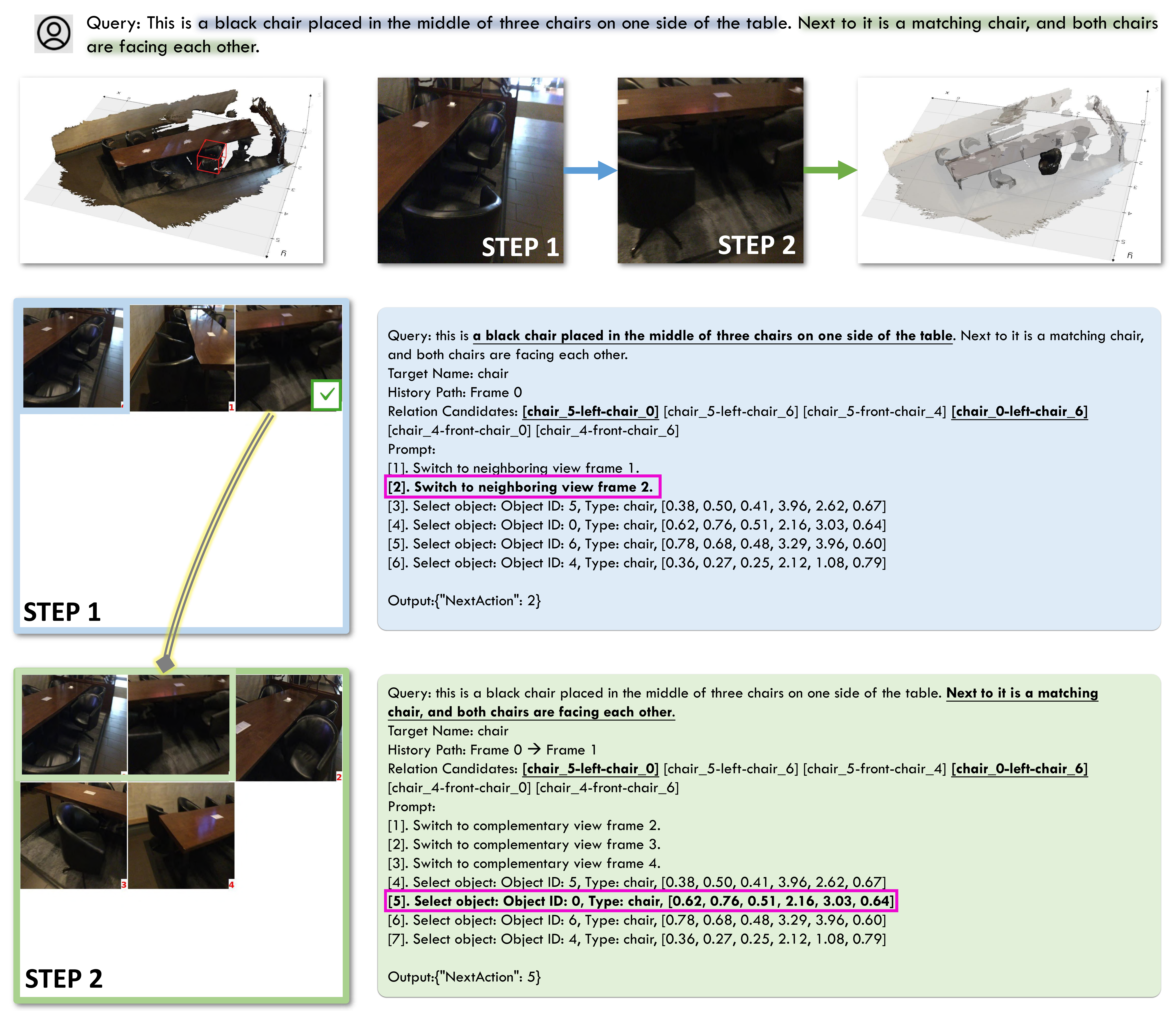}
\caption{
An example reasoning candidates of VoG framework for 3DVG.
}
\label{fig:candidates}
\end{figure*}

\begin{table*}[t]
  \centering
  \small
  \setlength{\tabcolsep}{5pt}
  \begin{tabular}{ll c  cc  cc  cc}
    \toprule
    \multirow{2}{*}{\textbf{Method}} 
    & \multirow{2}{*}{\textbf{Venue}}
    & \multirow{2}{*}{\textbf{Agent}}
    & \multicolumn{2}{c}{\textbf{Unique}}
    & \multicolumn{2}{c}{\textbf{Multiple}}
    & \multicolumn{2}{c}{\textbf{Overall}} \\
  \cmidrule(lr){4-5}\cmidrule(lr){6-7}\cmidrule(lr){8-9}
      &         &             
    & \textbf{Acc@0.25} & \textbf{Acc@0.5} 
    & \textbf{Acc@0.25} & \textbf{Acc@0.5} 
    & \textbf{Acc@0.25} & \textbf{Acc@0.5} \\
    \midrule
    SeeGround        & CVPR'25 & llava-v1.6-7B   &  —   &  —   &  —   &  — &  —  &  — \\
    \textbf{VoG}            & \textbf{Ours}    & llava-v1.6-7B   &  9.0   &  7.7   &  12.3   &  10.9 &  11.5  &  10.2 \\
    \midrule
    
    SeeGround          & CVPR'25 & DeepSeek-VL-1.3B   &  8.4   &  8.3   &  4.8   &  4.3 &  5.7  &  5.3 \\
    \textbf{VoG}            & \textbf{Ours}    & DeepSeek-VL-1.3B  &  
      47.9 \textcolor{red}{\scriptsize($\uparrow$39.5)} & 
      43.9 \textcolor{red}{\scriptsize($\uparrow$35.6)} &  
      22.2 \textcolor{red}{\scriptsize($\uparrow$17.4)} &  
      19.6 \textcolor{red}{\scriptsize($\uparrow$15.3)} &  
      28.4 \textcolor{red}{\scriptsize($\uparrow$22.7)} &  
      22.5 \textcolor{red}{\scriptsize($\uparrow$17.2)} \\
    \midrule

    SeeGround          & CVPR'25 & Qwen2.5-VL-3B      &  59.8   &  54.5   &  20.5   &  17.9 &  30.0  &  26.8 \\
    \textbf{VoG}            & \textbf{Ours}    & Qwen2.5-VL-3B   &  
      64.6 \textcolor{red}{\scriptsize($\uparrow$4.8)} &  
      58.7 \textcolor{red}{\scriptsize($\uparrow$4.2)} &  
      26.0 \textcolor{red}{\scriptsize($\uparrow$5.5)} &  
      22.6 \textcolor{red}{\scriptsize($\uparrow$4.7)} &  
      35.3 \textcolor{red}{\scriptsize($\uparrow$5.3)} &  
      31.3 \textcolor{red}{\scriptsize($\uparrow$4.5)} \\
    \midrule
    
    SeeGround    & CVPR'25 & Qwen2-VL-2B   & 59.9 & 55.0 & 20.7 & 18.4 & 30.2 & 27.2 \\
    \textbf{VoG}            & \textbf{Ours}    & Qwen2-VL-2B   &  
      69.1 \textcolor{red}{\scriptsize($\uparrow$9.2)} &  
      63.1 \textcolor{red}{\scriptsize($\uparrow$8.1)} &  
      25.3 \textcolor{red}{\scriptsize($\uparrow$4.6)} &  
      21.9 \textcolor{red}{\scriptsize($\uparrow$3.5)} &  
      35.9 \textcolor{red}{\scriptsize($\uparrow$5.7)} &  
      31.9 \textcolor{red}{\scriptsize($\uparrow$4.7)} \\
    \midrule
    
    SeeGround      & CVPR'25 & Qwen2-VL-72B  & 75.7 & 68.9 & 34.0 & 30.0 & 44.1 & 39.4 \\
    \textbf{VoG}            & \textbf{Ours}    & Qwen2-VL-72B  &  
      78.6 \textcolor{red}{\scriptsize($\uparrow$2.9)} &  
      71.5 \textcolor{red}{\scriptsize($\uparrow$2.6)} &  
      34.0 \textcolor{red}{\scriptsize($\uparrow$0.0)} &  
      30.4 \textcolor{red}{\scriptsize($\uparrow$0.4)} &  
      44.8 \textcolor{red}{\scriptsize($\uparrow$0.7)} &  
      40.3 \textcolor{red}{\scriptsize($\uparrow$0.9)} \\
    \bottomrule
  \end{tabular}
  \caption{Evaluations of zero-shot 3DVG on \textit{ScanRefer} validation set. Results are reported for “Unique” (scenes with a single target object) and “Multiple” (scenes with distractors of the same class) subsets, along with overall performance. Red numbers indicate absolute improvement over SeeGround baseline.}
  \label{tab:vlm_comparison}
\end{table*}

\begin{figure*}[t]
\centering
\includegraphics[width=.95\textwidth]{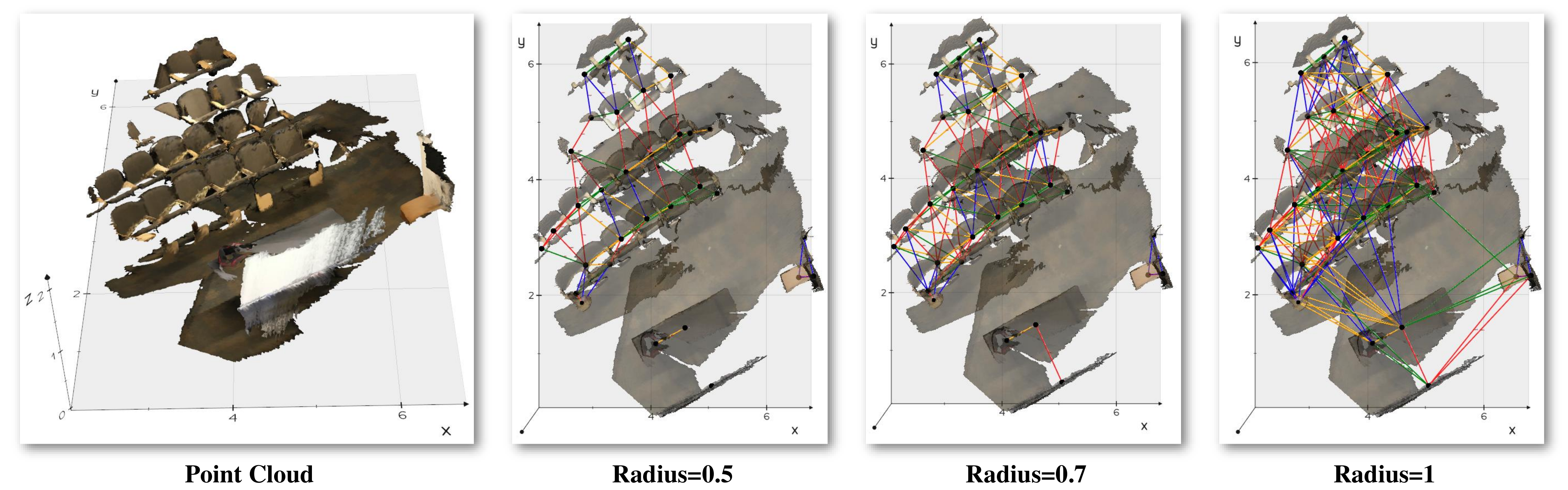}
\caption{
Graph radius analysis in object layer.
}
\label{fig:radius}
\end{figure*}

\section{Appendix B}

\subsection{VLM Prompt Design}

To operationalize the \textit{View-on-Graph} (VoG) paradigm in a practical multi-round reasoning loop, we design a structured prompt that explicitly constrains the Vision-Language Model (VLM) to follow the intended exploration and decision-making process.  
As illustrated in Fig.~\ref{fig:prompt}, the prompt consists of several key components, each aligned with VoG’s iterative traversal and reasoning mechanism:

\paragraph{1. Role Specification. }The prompt begins by assigning the VLM the role of a \emph{visual reasoning assistant} whose goal is to identify the object best matching the given natural language query in a 3D indoor scene. This framing ensures the model approaches the task as a step-by-step reasoning problem rather than a single-stage classification, matching VoG’s iterative exploration philosophy.

\paragraph{2. Environment Contextualization. } The prompt describes the scene as a reconstructed multi-view 3D environment, where each viewpoint is indexed as a \emph{frame} (e.g., \texttt{frame 4}). It specifies that each visible object has a unique \emph{Global Object ID} in the format \texttt{<class\_label>\_<index>} (e.g., \texttt{chair\_9}, \texttt{table\_16}), and that positions are estimated by a 3D detection model which may contain noise.  
    This context primes the VLM to reason under imperfect detection and encourages the integration of multiple viewpoints to reduce uncertainty, consistent with VoG’s motivation.

\paragraph{3. Task Format and Action Space.}  At each reasoning round, the prompt provides:  
    (1) a $3\times3$ grid image showing previously visited frames followed by the current frame,  
    (2) the natural language description of the target object, and  
    (3) a candidate action list consisting of:
    \begin{itemize}
        \item Switching to another frame (gather more context);
        \item Selecting an object (make final prediction).
    \end{itemize}  
    This discrete action space mirrors VoG’s two core operations: \emph{exploration} via graph traversal and \emph{reason} via grounding, ensuring that all inference steps are explicitly mapped to graph operations.

\paragraph{4. Interaction Constraints and Exploration Policy. } The prompt enforces a maximum of four reasoning rounds. Selecting an object ends the process immediately, preventing further exploration. Frame switching allows the VLM to continue exploring within the limit.  
    This directly corresponds to VoG’s bounded traversal depth $D_{\text{max}}$ and encourages strategic selection of viewpoints to maximize information gain per step.

\paragraph{5. Selection Strategy Guidelines.} The prompt instructs the VLM to \emph{switch frames when uncertain} and \emph{select an object when confident}.  
    This implements VoG’s decision policy: continue graph traversal until accumulated context satisfies the query constraints, then ground the object.

\paragraph{6. Output Format Enforcement.} The prompt requires a strict JSON output format:
\begin{verbatim}
{"NextAction": <number>}
\end{verbatim}
where \texttt{<number>} is the integer index extracted from the triple-square-bracket prefix in the chosen action.  
This ensures unambiguous parsing in automated pipelines and preserves compatibility with VoG’s iterative controller, which parses the action and updates the traversal state.

Overall, the prompt bridges the high-level design of VoG with the VLM’s step-by-step execution: role specification defines the problem scope, environment context primes reasoning under multi-view uncertainty, structured inputs and actions map directly to graph traversal and object grounding, and strict output constraints guarantee reproducibility and seamless integration with the VoG loop.

\subsection{VoG Candidates Generation}

Fig.~\ref{fig:candidates} illustrates an example multi-round reasoning process within our VoG framework for 3D visual grounding.  
Given the query \textit{"this is a black chair placed in the middle of three chairs on one side of the table. Next to it is a matching chair, and both chairs are facing each other"}, VoG starts from an initial viewpoint and incrementally explores the scene to locate the target.

At each reasoning step, VoG dynamically generates a \emph{candidate set} that drives the next decision. This candidate set directly corresponds to the available actions in VoG’s graph traversal space and consists of the following components:

\paragraph{1. View-switching candidates.}  Derived from the \emph{view--view edges} ($E^{VV}$) in the MMMG, these correspond to neighboring or complementary viewpoints that are directly connected to the current view node. 
Switching to one of these views allows the model to gather additional context about partially visible or occluded objects.  For example, in \textbf{Step~1}, the model receives two neighboring views (\texttt{[1], [2]}) as options to potentially improve visibility of the target chair.

\paragraph{2. Object-selection candidates.}  Derived from the \emph{view--object edges} ($E^{VO}$) of the current viewpoint, each entry specifies a visible object’s unique \emph{Global Object ID}, semantic type, and bounding box parameters (center coordinates and size).  
Selecting one of these objects constitutes a final grounding decision.  
For example, in \textbf{Step~2}, object \texttt{chair\_5} is presented with its spatial metadata \((0.38, 0.50, 0.41, 3.96, 2.62, 0.67)\), enabling the VLM to match it against the query.

\paragraph{3. Relation candidates.}  Extracted from the \emph{object--object edges} ($E^{OO}$) of the MMMG, these describe spatial relations between visible objects (e.g., \texttt{chair\_5-left-chair\_0}, \texttt{chair\_4-front-chair\_0}). These relations are injected into the prompt to guide the VLM’s reasoning, allowing it to verify query constraints such as ``next to it is a matching chair'' or ``in the middle of three chairs'' during exploration.

The interplay of these candidate types allows VoG to integrate local geometric context (relation candidates) with cross-view exploration (view-switching candidates) and final selection (object-selection candidates).  
In \textbf{Step~1}, partial evidence leads the model to select \texttt{NextAction: 2} to switch to a neighboring view.  
In \textbf{Step~2}, the improved visibility and confirmed spatial relations allow the model to select the correct target (\texttt{NextAction: 5}).  
This candidate generation process is repeated at each round, ensuring that exploration and grounding are jointly optimized in VoG’s multi-round reasoning loop.

\section{Appendix C}
\subsection{Different VLM Agents}

To further investigate how different VLM backbones influence the performance of our VoG framework, we evaluate multiple representative vision–language architectures (Table~\ref{tab:vlm_comparison}).  
We analyze the results in the context of each model’s architectural characteristics and the way VoG interacts with them.

\paragraph{LLaVA-v1.6-7B.}
LLaVA employs a CLIP-based vision encoder to produce a fixed set of visual tokens, which are concatenated with text tokens and fed into a LLaMA-based~\cite{touvron2023llama} decoder.  
The total sequence length is strictly bounded by the model’s maximum context window (e.g., 4096 tokens), meaning longer text segments reduce the number of retained visual tokens and vice versa.  
In complex 3DVG settings, both SeeGround and our method must encode an object list, scene description, and candidate actions, often pushing against this limit.  
For SeeGround, each scene is represented as a single rendered image with visual markers combined with a full list of detection box descriptions in text.  
Given LLaVA’s limited context length, this concatenated text frequently exceeds the model’s maximum input capacity, causing most detection box prompts to be truncated. As a result, crucial grounding cues are lost and the model fails to produce valid predictions.  

In contrast, VoG’s stepwise traversal restricts each reasoning round to a compact set of candidate actions, dramatically reducing prompt length while preserving critical scene information.  
This design allows more essential tokens to survive truncation and enables VoG to generate valid predictions in small or relatively simple reasoning scenarios.  
However, in more complex settings, such as when multiple instances of the same category appear in a single scene, the accumulated object pool after the second reasoning round can still exceed LLaVA’s readable length, causing similar truncation issues.  
Consequently, for LLaVA, our method yields valid outputs primarily in small-scale or low-complexity reasoning cases.

\paragraph{DeepSeek-VL-1.3B.}
DeepSeek is a lightweight, efficiency-focused multimodal model with a small language backbone and a modest vision encoder.  
Its limited long-form reasoning capacity and weaker fine-grained vision–language alignment make it more error-prone in challenging cases, particularly in the \emph{Multiple} subset where distractors share high visual similarity.  
Consequently, SeeGround’s performance with DeepSeek is low (5.7\% overall Acc@0.25).  
VoG mitigates these weaknesses by applying \emph{incremental retrieval}, only the most relevant visual–semantic cues are passed to the VLM at each step, reducing cognitive load and making the decision space more tractable.  
This targeted prompting significantly boosts DeepSeek’s grounding accuracy (5.7$\rightarrow$28.4 Acc@0.25), narrowing the gap with much larger VLMs.

\paragraph{Qwen2-VL Series.}
Qwen2-VL benefits from strong multimodal alignment and robust long-context handling, allowing it to process longer candidate lists and richer visual context without severe truncation.  
Even so, VoG still outperforms SeeGround across all Qwen2-VL variants.  
For the smaller Qwen2-VL-2B, VoG’s structured exploration yields notable gains in \emph{Multiple} scenes (20.7$\rightarrow$25.3 Acc@0.25), where contextual disambiguation is critical.  
For the large Qwen2-VL-72B, VoG achieves state-of-the-art zero-shot performance (44.8 Acc@0.25 overall), demonstrating that our traversal–reasoning paradigm remains beneficial even for high-capacity VLMs.

Overall, these results highlight that VoG’s design, compact per-step prompts, targeted retrieval, and graph-guided exploration, consistently benefits VLMs regardless of scale.  
For small models, it alleviates capacity limitations; for large models, it further enhances reasoning efficiency and accuracy.

\subsection{Graph Structure Analyses}

\begin{figure}[tbp]
    \centering
    \includegraphics[width=0.9\columnwidth]{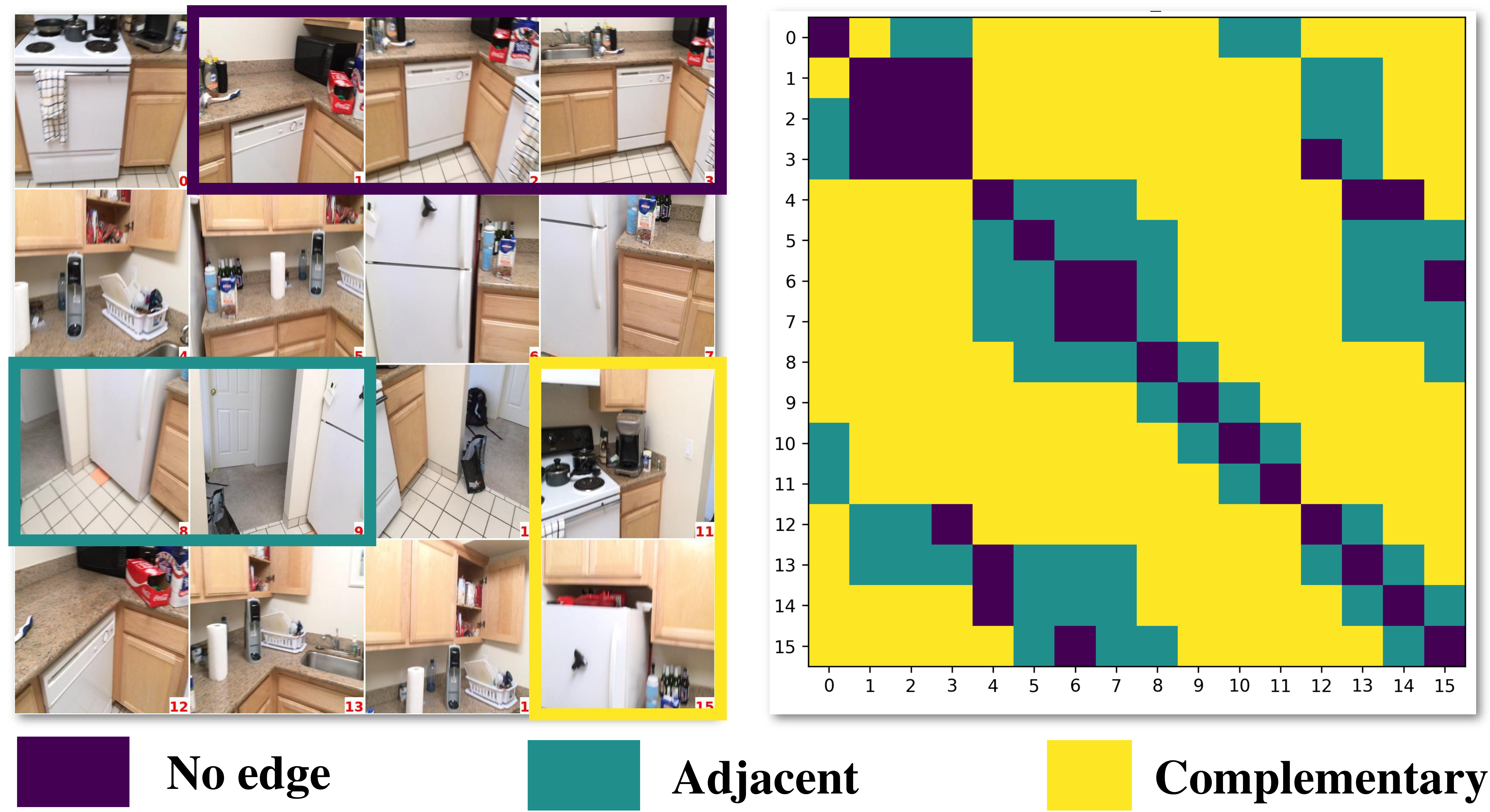}
    \caption{Edge type analysis in view layer.}

    \label{fig:edge_type}
    \end{figure}

The size of the MMMG is determined by the number of object nodes, the number of view nodes, and the density of their connecting edges.  
While the object nodes already cover all detected bounding box candidates, and the number of view nodes has been analyzed in the main paper, here we focus on discussing the density of the \emph{object–object edges} ($E^{OO}$) and the \emph{view–view edges} ($E^{VV}$).

\paragraph{1. Object–Object Edges ($E^{OO}$).}
The $E^{OO}$ edges are designed to capture \emph{short-range spatial relationships} that reflect the fundamental spatial layout of the scene.  
As shown in Fig.~\ref{fig:radius}, once the 3D scene is constructed into a graph, the spatial associations between objects become clearly visible.  
We visualize three cases using different connection radii: $r=0.5$, $r=0.7$, and $r=1.0$.  
A smaller radius yields a sparser graph that retains only immediate spatial neighbors, while a larger radius produces a denser graph including many long-range connections.  
Since long-range dependencies can be effectively captured through \emph{view–view} relations, we select $r=0.5$ for $E^{OO}$.  
This choice preserves a clear local spatial structure, avoids excessive edge density, and ensures that the generated candidate descriptions remain compact, preventing the VLM prompt length from becoming a burden during reasoning.

\paragraph{2. View–View Edges ($E^{VV}$).}
For $E^{VV}$, we define two types of edges: \emph{adjacent} and \emph{complementary}.  
Adjacent edges capture \emph{visual relatedness} between viewpoints, allowing the model to refine spatial reasoning by leveraging overlapping or partially similar views.  
Complementary edges link viewpoints that cover largely disjoint areas of the scene, enabling the VLM to \emph{explore novel regions faster} and avoid wasting reasoning steps on redundant observations.  
As illustrated in Fig.~\ref{fig:edge_type}, for a scene with $16$ viewpoints, the adjacency matrix reveals that:
\begin{itemize}
    \item View pairs with \textbf{no edge} (purple) generally observe almost identical regions, providing little additional information.
    \item \textbf{Adjacent} view pairs (teal) share overlapping content but also contribute some new local information.
    \item \textbf{Complementary} view pairs (yellow) cover distinct parts of the scene, offering maximal exploration gain.
\end{itemize}
By combining adjacent and complementary edges, the graph structure ensures that the VLM can both maintain contextual continuity and expand its coverage of the scene within the limited reasoning depth.

Overall, controlling the density of $E^{OO}$ and $E^{VV}$ allows us to maintain a compact yet expressive MMMG representation.  
This design preserves essential spatial–semantic structure while keeping the candidates size manageable for the VLM, ensuring efficient multi-round reasoning.

\begin{figure*}[t]
\centering
\includegraphics[width=.95\textwidth]{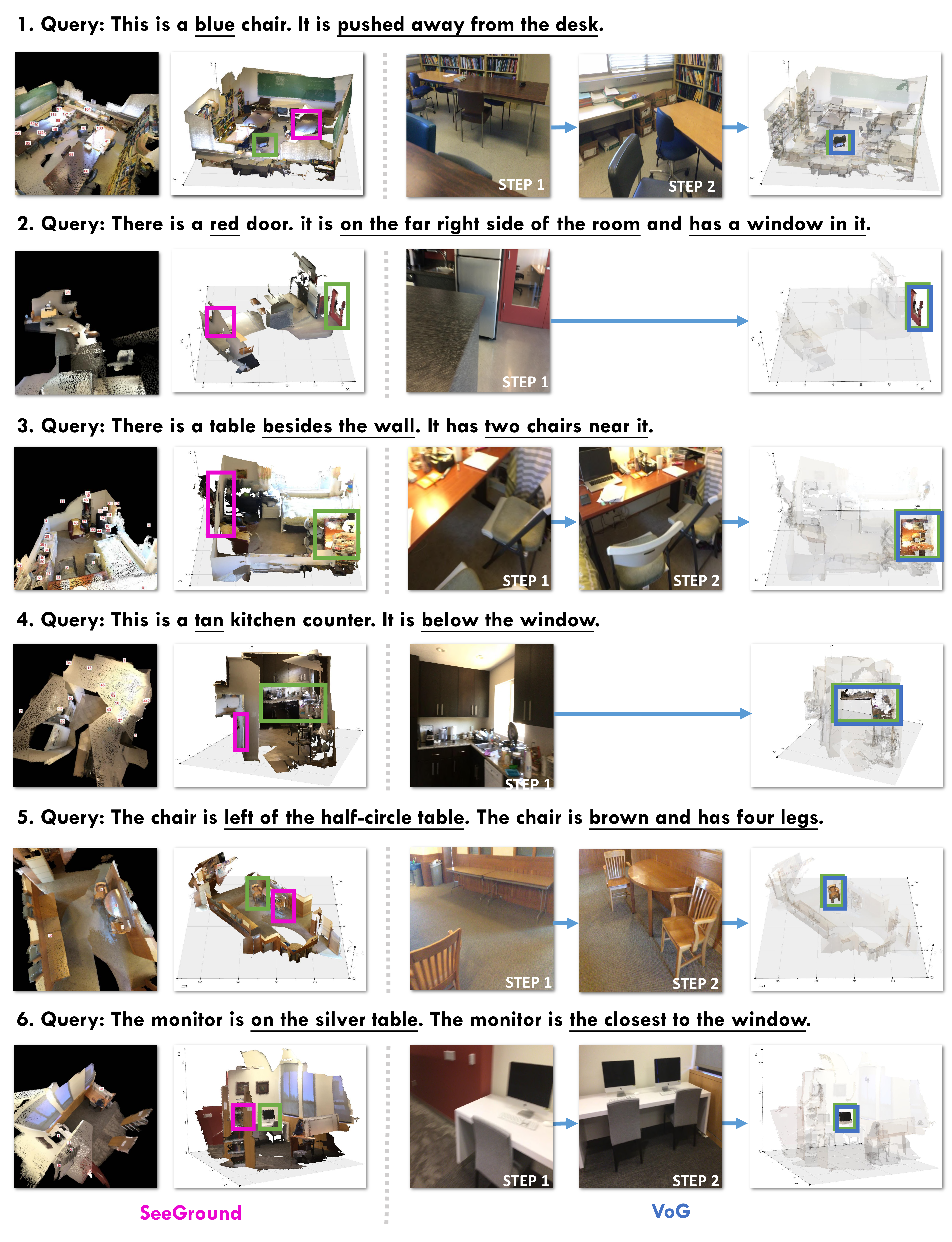}
\caption{\textbf{Qualitative grounding results.} Comparison between the existing state-of-the-art method SeeGround (left) and our method, VoG (right). The green box indicates the ground truth target object, the purple box shows SeeGround's grounding result, and the blue box highlights our grounding result using VoG.}
\label{fig:compare}
\end{figure*}

\begin{figure*}[t]
\centering
\includegraphics[width=.95\textwidth]{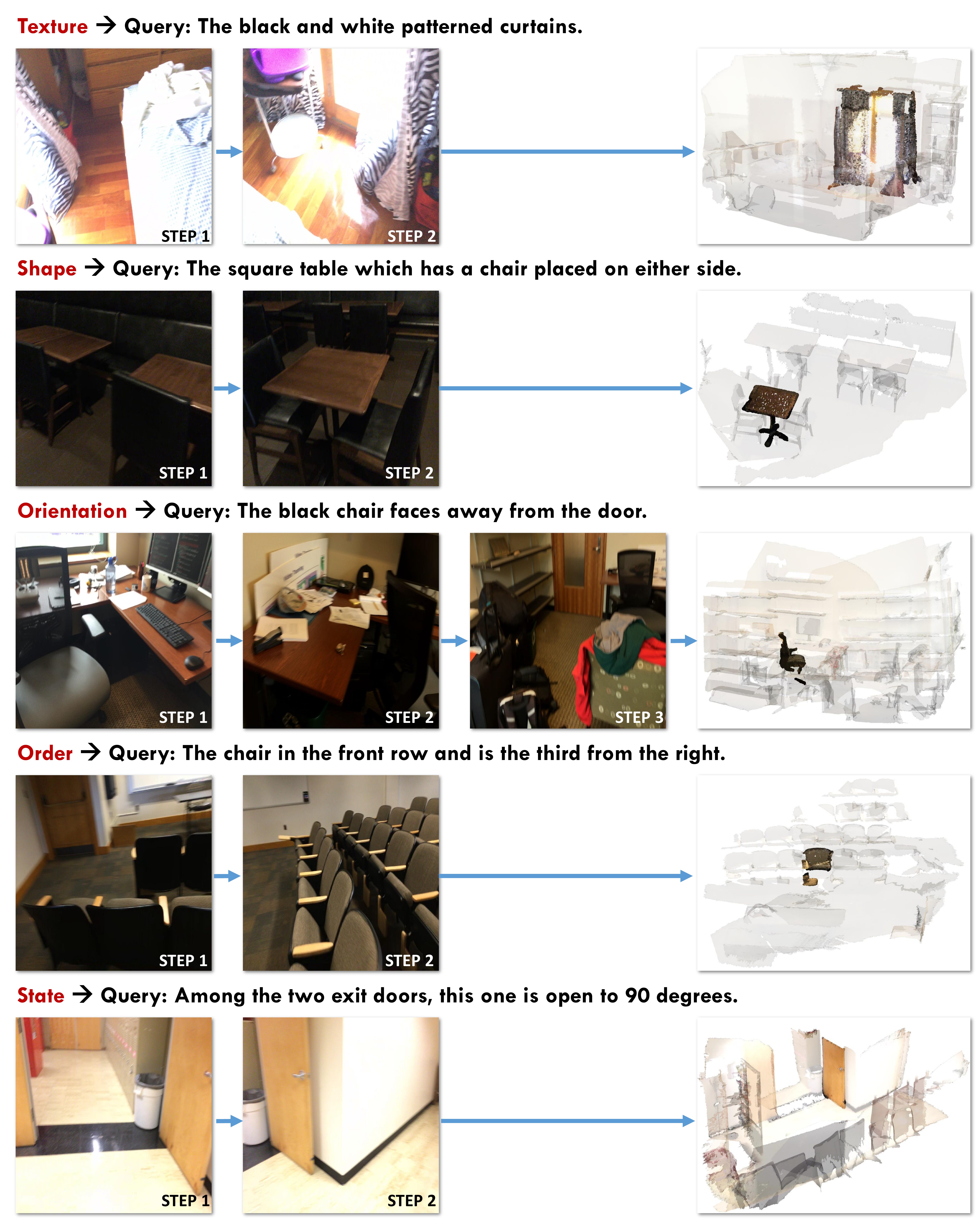}
\caption{\textbf{Qualitative grounding results.} The results of VoG method from different perspectives. Each query addresses different aspects, such as texture, shape, orientation, order, and state.}
\label{fig:attributes}
\end{figure*}

\section{Appendix D}
\subsection{Qualitative Grounding Results} 
As shown in Fig.~\ref{fig:compare}, we provide additional comparisons between the passive, composite visual input strategy of SeeGround~\cite{li2025seeground} and our active, external graph traversal exploration in VoG.  
SeeGround renders a single view centered on \emph{anchor} objects parsed from the text description, determined from detected bounding box labels in the scene.  
This strategy inherently depends on the accuracy of (i) text parsing to identify anchor and target objects, and (ii) the anchor-based view rendering to capture sufficient visual evidence.  
Failures in either stage can mislead the grounding process, even for visually simple \emph{Unique} scenes.

\begin{itemize}
    \item \textbf{Multiple instances of the same class (Row 1).}  
    The query describes a \emph{blue chair} in relation to a desk, but the scene contains multiple blue chairs of similar appearance.  
    SeeGround centers the rendered view on a suboptimal chair, leading to confusion and incorrect grounding.  
    VoG, in contrast, begins from an initial observation and iteratively explores alternative viewpoints, verifying spatial relations until it isolates the correct chair.

    \item \textbf{Parsing or localization errors in simple scenes (Rows 2 and 4).} 
    Even when the target object is unique in the scene (e.g., \emph{red door with a window}, \emph{tan kitchen counter below the window}), parsing errors or poor viewpoint selection in SeeGround prevent reliable grounding.  
    VoG’s stepwise reasoning progressively confirms object attributes and relations, leading to robust localization.

    \item \textbf{Loss of fine-grained spatial details (Rows 3 and 5).}  
    Some queries require fine spatial cues (e.g., \emph{table beside the wall with two chairs near it}, \emph{monitor closest to the window}).  
    Although SeeGround may capture the relevant objects in one view, long viewing distance and lack of detail hinder accurate reasoning.  
    VoG navigates to closer and more informative views, verifying positional relations and selecting the correct target.
\end{itemize}

Overall, these examples demonstrate that VoG’s active, external graph traversal mitigates the key weaknesses of SeeGround: dependence on perfect parsing, vulnerability to occlusion, and loss of fine spatial detail in single views.

In Fig.~\ref{fig:attributes}, we also showcase the effectiveness of the VoG method in handling diverse queries by visualizing the results from different perspectives, such as texture, shape, orientation, order, and state.

\section{Appendix E}
\paragraph{Future Work.}
The concept of employing an external structured graph and treating a large model as an \emph{agent} to iteratively traverse it has been widely explored in the NLP domain.  
Existing Knowledge Graph (KG) based reasoning frameworks can be grouped into three categories:  
(1) \textbf{Direct LLM reasoning on text.} The model answers questions based solely on its parametric memory, without explicit external retrieval, relying on prompt engineering or chain-of-thought strategies to simulate multi-hop reasoning.  
(2) \textbf{Loose coupling between LLM and KG.} The model first converts a natural language query into a formal symbolic query or relation sequence, retrieves facts from the KG, and then synthesizes an answer. In this setting, the LLM does not directly participate in the graph traversal itself.  
(3) \textbf{Tight coupling between LLM and KG.} The model acts as an agent that actively explores the KG in multiple steps, selecting relevant entities and relations at each stage, and incrementally building reasoning paths until sufficient evidence is collected~\cite{sun2023think}.

These paradigms have proven effective in text-based knowledge reasoning, our viewpoint-guided 3D visual grounding task differs substantially in several aspects.  
First, \textbf{modality alignment} becomes a central challenge: the reasoning agent must bridge between textual queries and a multi-modal, visually grounded SG, requiring robust mapping from linguistic entities to detected visual objects.  
Second, \textbf{per-step inference cost} is significantly higher: in NLP, symbolic graph exploration incurs low computational cost, whereas in visual grounding, each step involves visual feature processing and multi-modal reasoning through a VLM, making deep or wide traversal more expensive.  
Third, the \textbf{graph structure itself is perceptually derived}: edges encode spatial–semantic relations and visual co-occurrence patterns, which may be incomplete or noisy due to perception errors, unlike curated KGs in NLP. In future work, we plan to \textbf{design cost-aware multi-modal traversal strategies} that dynamically select high-value viewpoint transitions while controlling VLM inference overhead, we aim to evolve VoG into a scalable, interpretable, and cost-efficient framework for complex 3D visual grounding.

\section{Acknowledgments}
This work was supported in part by National Natural Science Foundation of China under Grants
62441216 and 62502068.

\clearpage

\bibliography{aaai2026}

\end{document}